\documentclass[conference]{IEEEtran}
\usepackage{times}

\usepackage[numbers]{natbib}
\usepackage{multicol}
\usepackage[bookmarks=true]{hyperref}

\usepackage{graphicx}
\usepackage{amsmath}
\usepackage{amssymb}
\usepackage{siunitx}  
\usepackage{subcaption}
\usepackage{gensymb}
\usepackage{booktabs}
\usepackage[ruled,vlined,linesnumbered,noresetcount,longend]{algorithm2e}
\usepackage{xcolor}
\let\labelindent\relax 
\usepackage{enumitem}
\newcommand{\rulesep}{\unskip\ \vrule\ }

\begin{document}

\title{A Behavioral Approach to Visual Navigation\\with Graph Localization Networks}



\author{\authorblockN{Kevin Chen\authorrefmark{1},
Juan Pablo de Vicente\authorrefmark{2},
Gabriel Sep\'ulveda\authorrefmark{2}, 
Fei Xia\authorrefmark{1},\\
Alvaro Soto\authorrefmark{2},
Marynel V\'azquez\authorrefmark{3}, and
Silvio Savarese\authorrefmark{1}}
\authorblockA{\authorrefmark{1}Stanford University}
\authorblockA{\authorrefmark{2}Pontificia Universidad Cat\'olica de Chile}
\authorblockA{\authorrefmark{3}Yale University}}


%

\maketitle

\begin{abstract}
Inspired by research in psychology, we introduce a behavioral approach for visual navigation using topological maps. Our goal is to enable a robot to navigate from one location to another, relying only on its visual input and the topological map of the environment. We propose using graph neural networks for localizing the agent in the map, and decompose the action space into primitive behaviors implemented as convolutional or recurrent neural networks. Using the Gibson simulator, we verify that our approach outperforms relevant baselines and is able to navigate in both seen and unseen environments. \\
Webpage URL: \url{https://graphnav.stanford.edu}.


\end{abstract}

\IEEEpeerreviewmaketitle

\section{Introduction}

Despite the ever-changing state of our indoor environments due to rearrangements in furniture, changes in lighting, or the simple accumulation of clutter, humans are able to seamlessly navigate through these ``dynamic'' spaces as if nothing in the environment had changed at all. How can we build similar visual navigation systems for robots? Although most approaches for visual navigation today rely on metric maps of the world and precise localization \cite{thrun2005probabilistic}, research suggests that biological systems in mice and men rely on coarse spatial layout representations in the form of ``cognitive maps'' \cite{tolman1948cognitive}. Studies suggest that at the core of such cognitive maps for large-scale spaces is a topological description of the environment that can capture relationships about different locations \cite{kuipers1991robot,siegel1975development,lynch1960image,piaget56:book}. Animals then execute navigation strategies based on their qualitative knowledge of the space \cite{foo2005humans}. While topological representations for navigation have been extensively explored in the past, the re-emergence of neural networks as a powerful tool for solving a variety of tasks motivates us to revisit early ideas of topological robot navigation and behavioral control \cite{franz2000biomimetic,brooks1986robust}.

\begin{figure}
    \centering
    \includegraphics[width=0.75\linewidth]{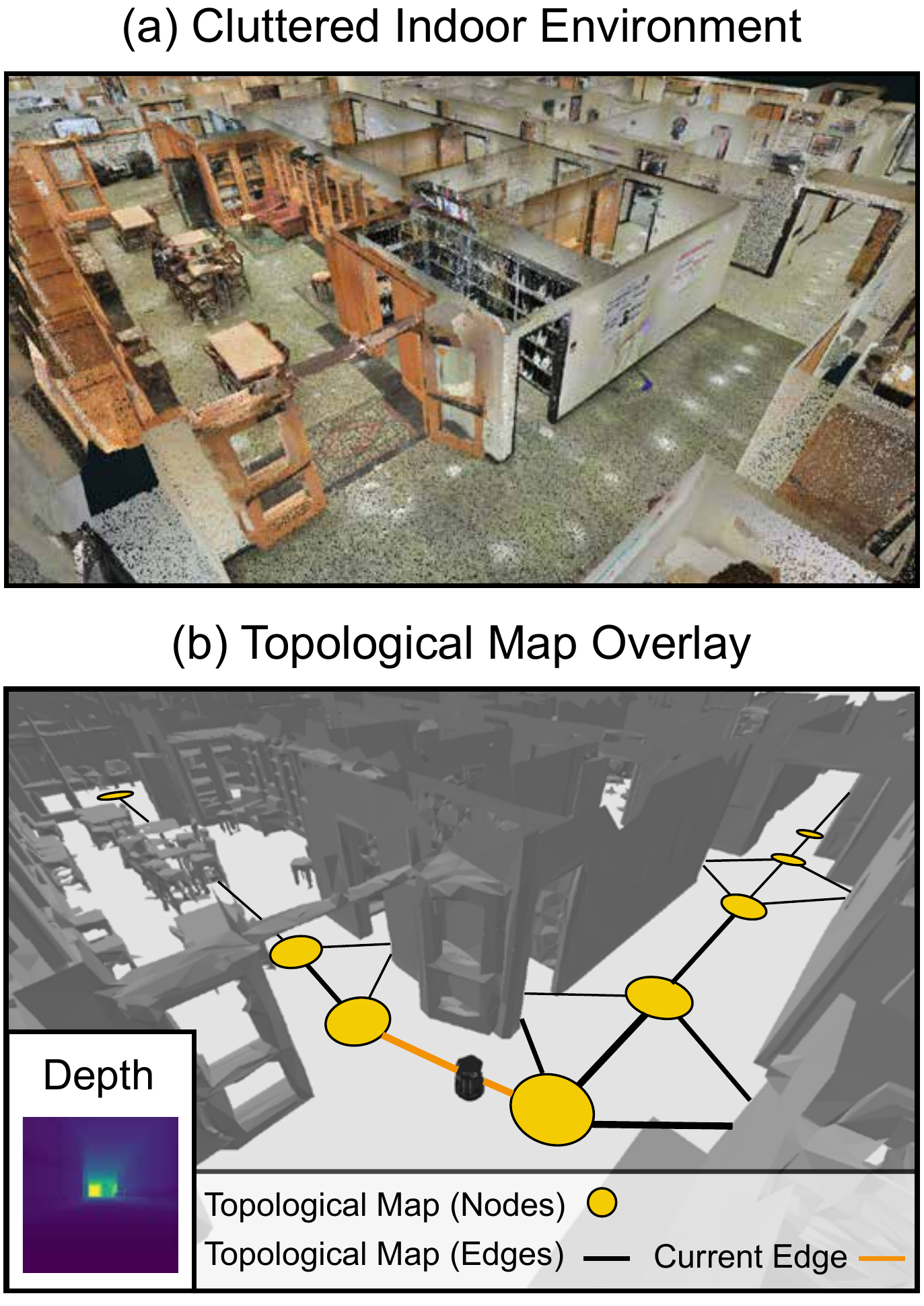}
    \vspace{-0.3em}
    \caption{A robot operates in a cluttered indoor environment (a). Using a topological representation of the environment and depth observations (b), the robot must navigate to its destination, specified as a location within the topological map.}
    \label{fig:pull}
    \vspace{-1em}
\end{figure}

In this work, our goal is to enable a robot to navigate from place A to place B given a topological map, a plan for how to get from A to B in the map, and the current observation of the environment obtained with an on-board camera (Figure \ref{fig:pull}). To tackle this navigation problem, we borrow ideas from both classical robotic architectures and newer deep learning approaches. In particular, we use a behavioral approach \cite{brooks1986robust,horswill1993polly} in conjunction with a topological representation of the environment, but leverage modern deep learning techniques to perform tasks such as localization and low-level control.


We pose the problem of robot navigation as a graph traversal problem in a topological representation of the environment. This formulation leads to three key questions: (1) What is an effective topological representation for navigation? (2) How do we localize the agent within this topological representation? And (3) given localization information, how do we control the robot to move according to our plan? 

To address question (1), we construct the map to be a directed graph with coarse information about relevant locations for the navigation task (nodes) and connectivity between close locations (edges). Every edge of the map is also labeled with a visuo-motor behavior, such as turn left or turn right, that can be executed to move from the source to the target node.

Given this representation of the environment, the next question is (2) how to localize the agent within the map. We propose using convolutional neural networks  and graph neural networks, which have been shown to have state-of-the-art performance in node classification and graph classification problems \cite{xu2018powerful}. 
Graph neural networks (GNN) are a natural approach for solving graph-related inference tasks because their architecture allows them to capture relational inductive biases, e.g., as specified by a topological map. We use these networks in our approach to infer the location of a robot based on the environment topology and its current visual input.

Lastly, given a localization estimate, the robot must determine how to maneuver itself according to the plan (3). By construction, a path in the topological map can be translated to a navigation plan in the form of a sequence of behaviors. It is therefore trivial to determine which behavior to execute given a localization prediction and a plan. 
We use neural networks to robustly execute the given behavior, and repeat localization and low-level control at every timestep (e.g. 5 Hz) to ensure smooth transitions to different parts of an environment.

We test our approach on the Stanford 2D-3D-S dataset \cite{armeni_cvpr16,armeni_arxiv_2d3ds}, which consists of reconstructed meshes of several university buildings with complex layouts and large amounts of clutter. We first constructed and annotated maps of these environments with the proposed topological descriptions. We then incorporated the maps into the Gibson physics-based simulator \cite{xiazamirhe2018gibsonenv} and extended the simulator's capabilities to create a testbed for benchmarking robot navigation approaches. We contribute to the community our map specification as well as our tools and data, including a dataset that we created for training the learning components of our system. This dataset is composed of 2,371 long sequences of robot observations (e.g., RGB, depth, and semantic data) of the 2D-3D-S buildings and the corresponding robot locations in topological maps. 
Using this setup, we show that our method can  navigate more robustly than relevant baselines in both seen and unseen cluttered indoor environments.

In summary, the main contributions of our work are:
\begin{itemize}[wide, labelwidth=!, labelindent=0pt,noitemsep]
    \item We introduce a specification for topological map design in complex, real-world environments.
    \item We propose a novel framework for localization and navigation using convolutional neural networks in conjunction with graph neural networks. By using a behavioral approach, we are able to robustly navigate through realistic environments using behavior-specific neural networks.
    \item We provide a dataset of robot navigation trajectories with corresponding topological maps. Along with the trajectories, we share RGB, depth, and semantic observations, as well as annotated node localization and edge behavior information. 
    \item We provide a testbed for benchmarking navigation tasks with topological maps using  Gibson. Agents can be controlled through the Robot Operating System (ROS), and our evaluation suite allows us to thoroughly analyze their performances.
\end{itemize}

\section{Related Work}

There is a long history of prior work on indoor robot navigation. We refer the reader to \cite{Kostavelis:Gasteratos:2015,thrun2005probabilistic} for a comprehensive overview of related work. Due to limited space, we focus our discussion on recent works related to visual indoor navigation using deep learning (DL).

In contrast to the classical approach to robot navigation that separates the tasks of mapping and localization \cite{thrun2005probabilistic} from the task of path planning
\cite{Latombe:Planning:1991}, most recent DL based models propose end-to-end solutions that directly map sensor data to robot actions \cite{zhu2017target,dosovitskiy2016learning,gupta2017cognitive,savinov2018semi}. In a pioneering work, \citet{zhu2017target} presents a deep reinforcement learning (DRL) approach that uses visual inputs to learn to steer a robot to a goal position without resorting to an explicit map of the environment. Interestingly, the resulting action policy learns navigational priors that depend on the contextual scene, e.g. kitchen or bedroom. As a drawback, due to its reactive nature, the approach exhibits limited capabilities to generalize to new environments or to plan long trajectories. \citet{gupta2017cognitive} use DL techniques to emulate the Bayesian cycle behind SLAM techniques \cite{thrun2005probabilistic}. However, their approach relies on ground truth ego-motion as input and operates in a discretized domain. 

In terms of works that use DL to keep an explicit representation of the environment, 
\citet{parisotto2017neural} present a Neural Map that emulates the operation of a 2D occupancy grid \cite{Elfes:1989}. As a main novelty, each grid cell stores an embedding that encodes a memory related to the corresponding robot position. Also in the context of a metric map, \citet{mirowski2016learning} propose an end-to-end DRL strategy that improves robot navigation by considering auxiliary tasks, such as depth prediction and loop closure classification. More similar to our approach, \citet{savinov2018semi} implements a topological representation where graph edges associate neighboring visual memories. Using this graph, robot localization is performed using a nearest neighbors approach.

Departing from the works above, we follow the traditional approach to robot navigation separating mapping and localization from planning and execution. Furthermore, using ideas from behavioral-based approaches \cite{brooks1986robust, horswill1993polly}, we use DL to implement a set of navigational behaviors such as \textit{follow a corridor} or \textit{turn right} \cite{pomerleau1989alvinn, bojarski2016end}. Our aim is to take full advantage of the rich semantic structure behind man-made environments for navigation, and leverage simple robot behaviors to perform complex tasks. Also, we avoid reliance on metric information \cite{parisotto2017neural}, robot odometry \cite{gupta2017cognitive}, or specific poses that need to be attained during indoor navigation \cite{gupta2017unifying}.

In terms of our topological representation, our work is closely related to that of \citet{sepulveda2018deep}. However, we do not rely on modifying the environment by introducing artificial landmarks, and we define a reduced set of primitive behaviors. These modifications help to (1) facilitate the design of topological maps for realistic, human environments, and (2) increase the robustness of learned navigation behaviors given limited data. Furthermore, the work in \cite{sepulveda2018deep} does not pose navigation as a graph traversal problem. 

There is also a connection between our work and previous efforts on compositionality 
\cite{andreas2017modular,andreas2016neural} and learning sequencing of manipulation primitives \cite{felip2013:article, huang2018neural,yu2018one}. However, our goal is not to solve new tasks given demonstrations of an ordered set of actions. Instead, our goal is to execute an abstract navigation plan given a topological map of a realistic human environment. 

A key component of our navigation approach is a localization network that leverages GNNs \cite{scarselli2009graph,battaglia2018relational}. Recently, GNNs have gained great attention as a powerful tool to model relational data \cite{xu2018powerful}. In terms of robot navigation, \citet{yang2018visual} uses graph convolutional networks \cite{kipf2016semi} but in the context of encoding semantic scene priors. As far as we know, we are the first to use GNNs to pose robot navigation as a graph traversal problem in a topological map of the environment. 

We test our approach using Gibson, an  environment for real-world perception \cite{xiazamirhe2018gibsonenv} of large-scale indoor spaces \cite{armeni_arxiv_2d3ds}. In particular, we create topological maps for different  spaces and extend the capabilities of Gibson to create a benchmark for behavioral robot navigation. With this effort, we support others, e.g., \cite{wu2018building, Dosovitskiy17, savva2017minos}, in building a rich and extensible environment for mobile robotics and machine learning research. 

\section{Problem Setup}

We consider a robot operating in a cluttered indoor environment with the goal of navigating from one node (A) in the topological map to another node (B). In our setup, the agent may not have seen the environment before, so no prior visual information is provided in relation to the map. The ground truth node location A is given to the robot when navigation begins, but it must rely on its visual input and the map to reach the desired destination.  It is crucial for the robot to avoid obstacles -- otherwise it will fail the navigation task.

To mimic realistic physical settings, we consider a ROS-controlled Turtlebot robot navigating in the PyBullet\cite{coumans2018}-powered Gibson simulator \cite{xiazamirhe2018gibsonenv} (Fig. \ref{fig:pull}b). We perform experiments with this robot on environments from the Stanford 2D-3D-S dataset \cite{armeni_cvpr16,armeni_arxiv_2d3ds}. The dataset was created  using a Matterport scanner to capture the geometry of office spaces in three different university buildings. As a result, the floor plans can be very complex and the spaces are filled with clutter including chairs, couches, tables, boxes, and even dollies.


\section{Topological Map Design}

Inspired by research in psychology \cite{kuipers1991robot,siegel1975development,lynch1960image,piaget56:book}, we use a topological representation -- a graph -- to encode spatial information about the environment. 
At a high level, each node in the topological map represents a location. Each edge corresponds to a behavior that allows the robot to get from the corresponding source node to the target node, similar to \citet{sepulveda2018deep}. However, due to the layout complexity of the naturalistic environments in the Stanford 2D-3D-S dataset, we substantially changed the topological map design compared to \citet{sepulveda2018deep}. In our work, the map annotations are done manually as described in the next paragraphs. Automating this process is a valuable future research direction \cite{savinov2018semi}. 

\begin{figure}
    \begin{subfigure}{0.35\linewidth}
    \includegraphics[width=\linewidth]{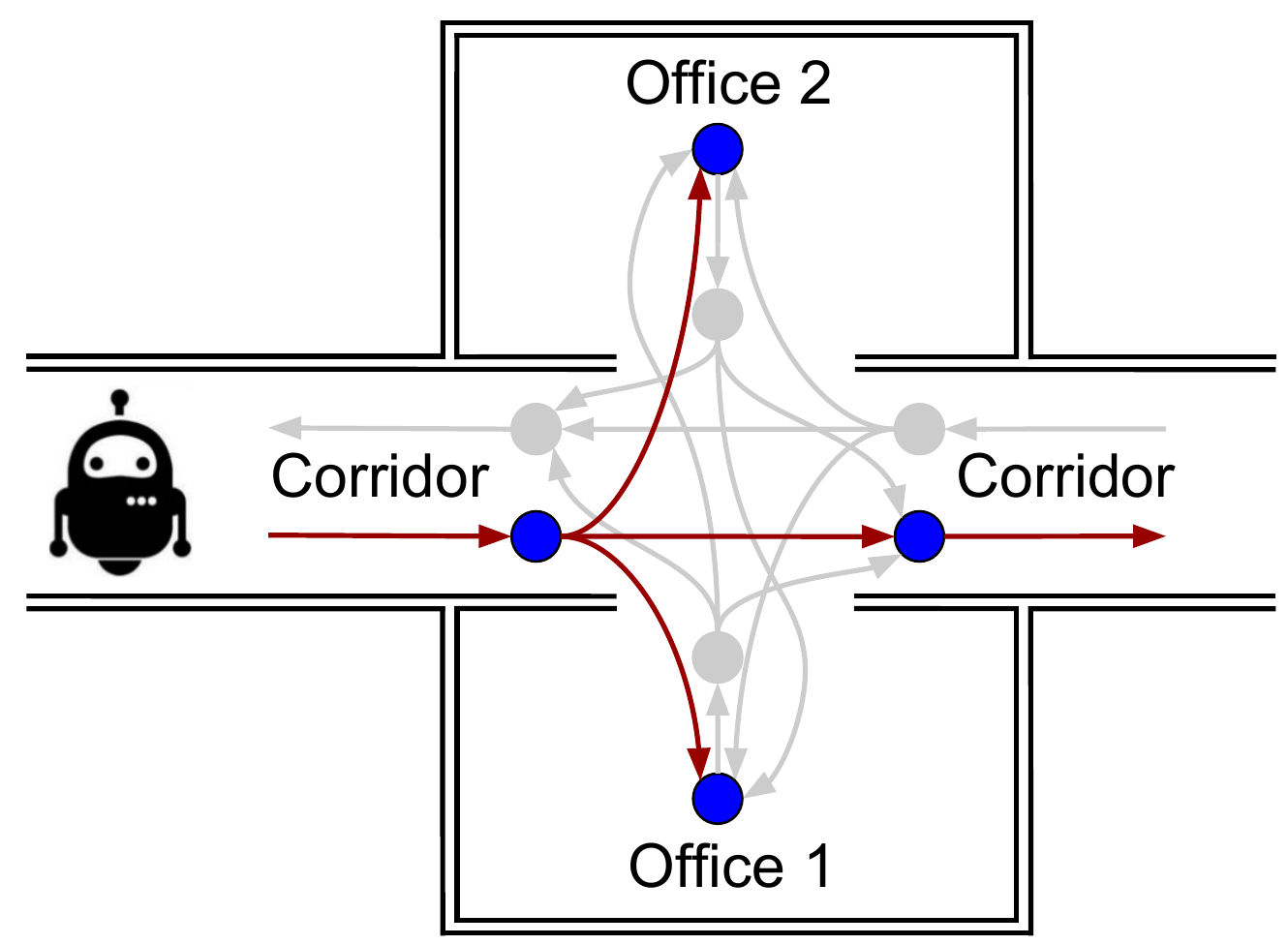}
    \caption{Entering a room}
    \label{fig:toy-topological-map-room-enter}
    \end{subfigure}
    \rulesep
    \begin{subfigure}{0.3\linewidth}
    \includegraphics[width=\linewidth]{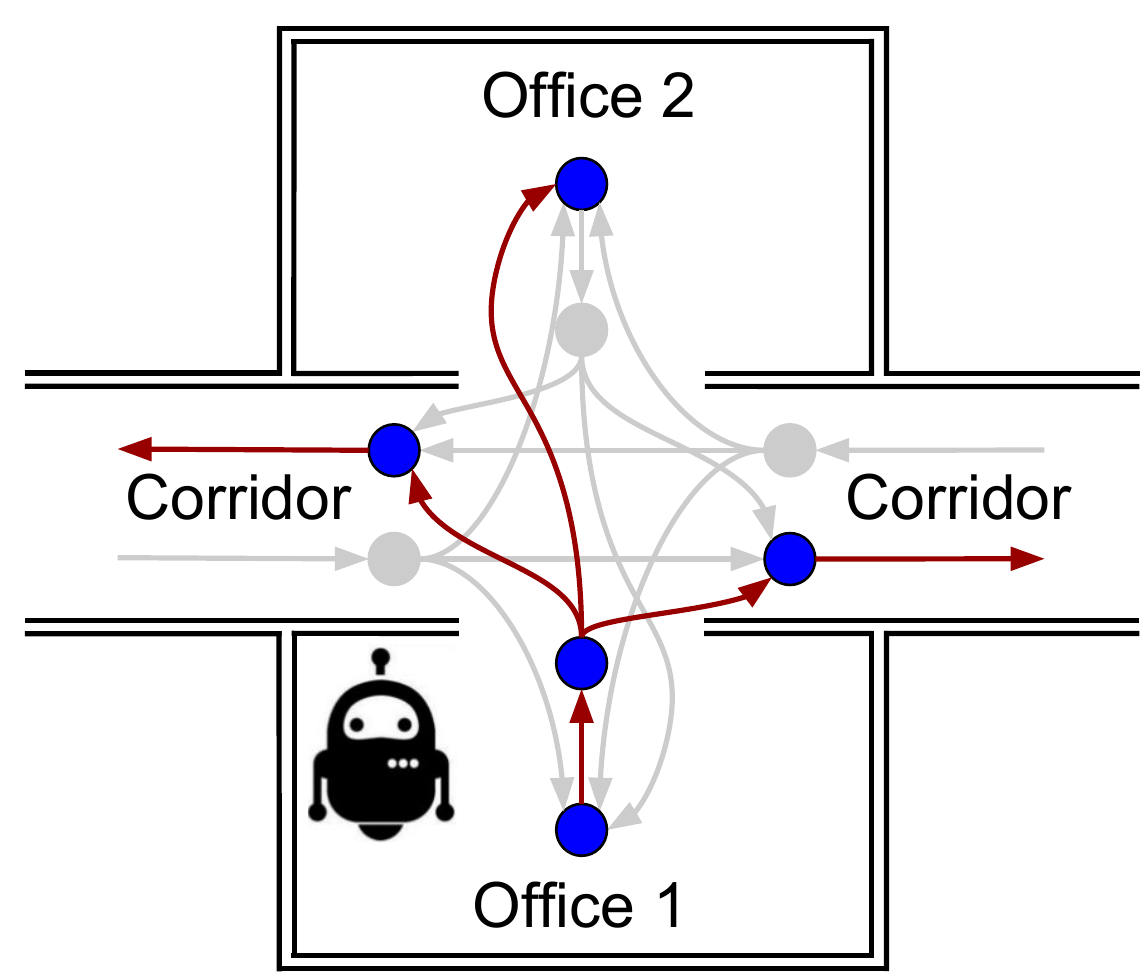}
    \caption{Exiting a room}
    \label{fig:toy-topological-map-room-exit}
    \end{subfigure}
    \rulesep
    \begin{subfigure}{0.25\linewidth}
    \includegraphics[width=\linewidth]{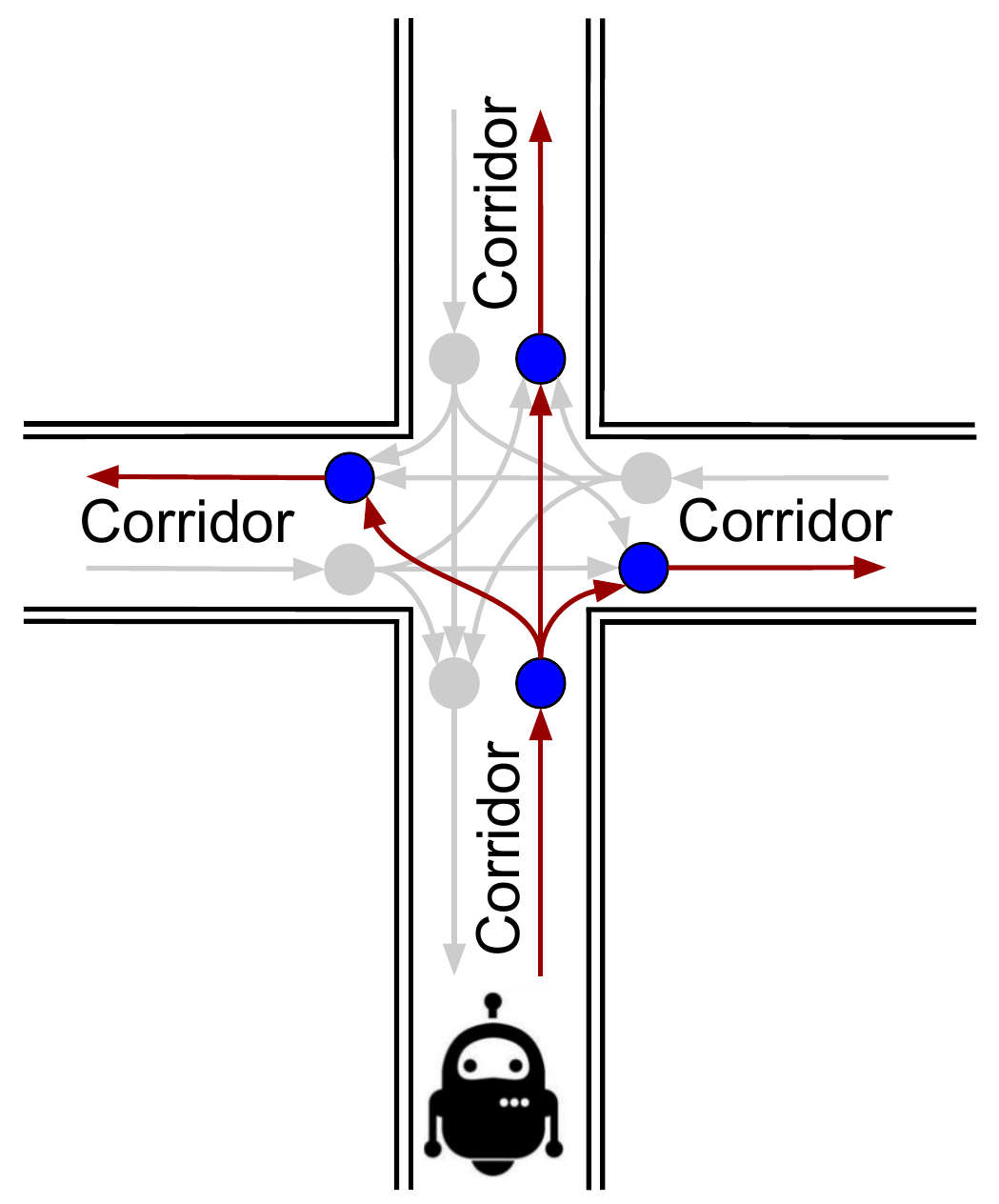}
    \caption{Entering an intersection}
    \label{fig:toy-topological-map-intersection}
    \end{subfigure}
    \caption{Examples of the topological map cropped to the local region of the agent.}
    \label{fig:toy-topological-map}
    \vspace{-2em}
\end{figure}

Our insight behind map design is that the robot should be able to traverse to and from any semantic location (e.g. office 1, office 2, pantry 1, conference room 1, etc.) by composing a \textit{minimal length} sequence of behaviors (edges) as specified by the topological map. We leverage the Manhattan world structure of indoor environments \cite{coughlan1999manhattan} and define the behaviors: \textit{\{find door (fd), corridor follow (cf), turn left (tl), turn right (tr), straight into room (s)\}}. 
We reduce the specificity of our behaviors (compared to \citet{sepulveda2018deep}) in order to simplify the design of topological maps, and make navigation more generalizable and transferable across different scenarios. Intuitively, turning left out of an office should require very similar controls to turning left at a four-way intersection/junction or turning left into a room (Fig.~\ref{fig:toy-topological-map-room-enter}).

To avoid localization ambiguities at each position in the topological map, nodes and edges also have associated orientations. For example, if executing a \textit{corridor follow} behavior in a hallway, it is not clear which direction to travel in, since an agent can move along two opposite directions. To resolve this ambiguity, there are two sets of nodes and edges for each corridor, one for each direction (Fig.~\ref{fig:toy-topological-map-intersection}). 

Similarly, each room has a corresponding room node that indicates that the robot is within the boundaries of that space and facing any direction. In this case, if the robot is facing towards the inside of the room, a turn behavior (e.g. turn left out of the room) is not well defined. Thus, to ensure smooth transitions in and out of these enclosed spaces, we also add door nodes to rooms. These door nodes indicate that the robot is positioned at the door and oriented towards the exit of the room, as shown in Fig.~\ref{fig:toy-topological-map-room-exit}.


Based on these observations, we formulate the topological map as a directed graph. We apply this representation to real world environments using the following rules:
\begin{enumerate}
    \item Each room in the environment, such as an office or conference room, has its own single node.
    \item Each room door also has its own node.
    \item The \textit{find door} behavior should connect each room node to its door node.
    \item Corridors have two sets of nodes, one for each direction of the corridor.
    \item Edges which indicate entering a room should connect to the room node.
    \item Exiting a room occurs from the door node.
\end{enumerate}
In general, nodes should be placed at any transition point -- that is, any location that may require a change of behavior. For example, upon approaching the exit to a room, there are many possible behaviors to execute, such as \textit{turn left}, \textit{turn right}, or even go \textit{straight} across the hallway into another room. Because this would require a change from the previous behavior (\textit{find door}), a door node should be placed prior to the exit of each room. Likewise, after an agent has turned into a hallway, the robot will likely transition from the turning behavior to a different behavior (e.g. \textit{corridor follow}). Therefore, a node must be placed immediately after the turn to signify the transition. Using this topological map representation, it is then trivial to compute the sequence of behaviors for navigating to and from any location (node) in the map with classical planning algorithms \cite{russell2016artificial}.


\begin{figure*}[t]
    \centering
    \includegraphics[width=0.78\linewidth]{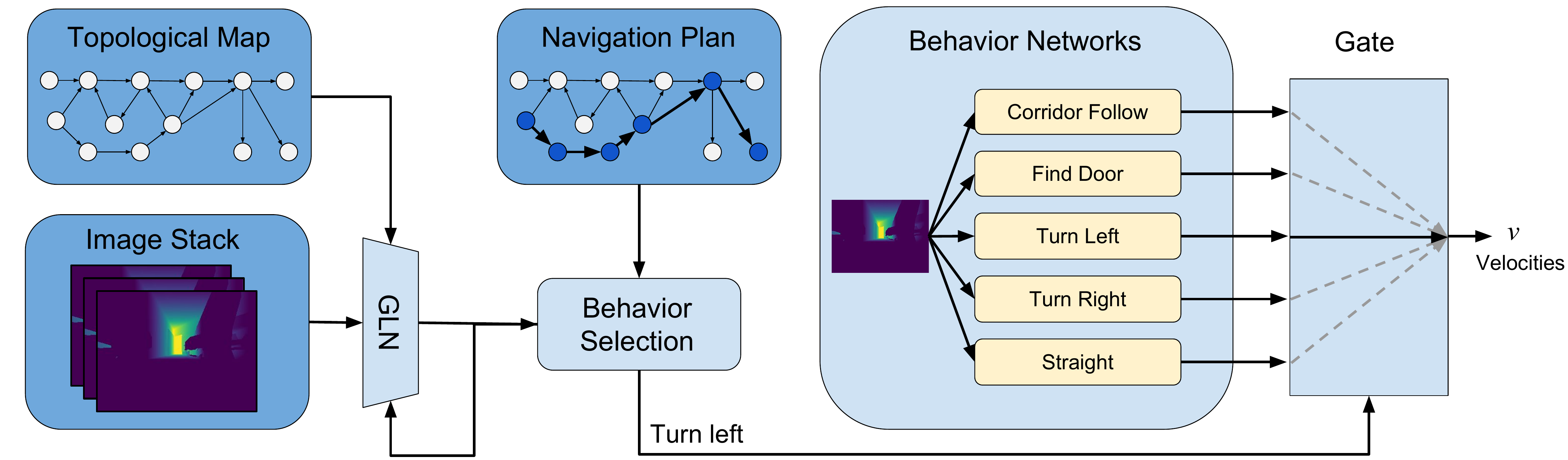}
    \caption{Our navigation approach addresses localization and behavior selection. Based on the localization estimate from the graph localization network (GLN) and the navigation plan, the agent can select a behavior network for the current timestep. The velocities output from the selected network are used for low-level motor control.}
    \label{fig:model-overview}
    \vspace{-1em}
\end{figure*}

\section{Method}


Two challenges for effective navigation  with our topological representation are: how to localize the agent, and how to direct the agent along a plan. We organize our navigation approach based on these key problems, as illustrated in Fig.~\ref{fig:model-overview}. 

The first challenge, localization, only needs to be done relative to the topological map. To this end, we propose using a graph neural network in combination with a convolutional neural network (CNN).
Once localized, the agent can easily plan paths to any destination in the environment. Moreover, it is trivial to determine which behavior needs to be executed given a localization prediction and a planned path.

The second challenge comes in during the execution of the plan. While the agent continuously updates its localization prediction, it also has to generate motion controls to ensure that it continues to follow the planned trajectory. In our work, for \textit{each} behavior we implement an individual neural network, which we call a behavior network. This kind of network takes as input the current observation of the world from the robot -- in our case, a depth image -- and predicts low-level velocity commands for executing the behavior.

The above components are combined using a simple behavior selection module, which looks up the correct behavior to execute given the localization prediction and the navigation plan. In the case that a localization estimate is not part of the plan, the module continues selecting the behavior from the last valid position that was part of the plan. Localization and behavior  selection are repeated 
until the agent reaches its destination or deviates from the expected path. 

The following Section provides a brief introduction to graph neural networks (Sec.~\ref{sec:gnn_overview}),
followed by a description of the graph representation used in our model (Sec.~\ref{sec:graph_representation}). Then, we describe our graph localization network (Sec.~\ref{sec:graph_localization_network}) and, lastly, present the behavior networks (Sec.~\ref{sec:behaviornet}).

\subsection{Preliminaries on Graph Neural Networks}
\label{sec:gnn_overview}

First introduced by \citet{scarselli2009graph}, GNNs have been shown to be very effective at learning relative inductive biases specified by graph structures. The following overview borrows heavily from the description and notation in \citet{battaglia2018relational}. For more details, we refer interested readers to \cite{battaglia2018relational}.

We define a directed graph to be a tuple $G = (\mathbf{u}, V, E)$, where $\mathbf{u}$ is a global feature for the graph and can be interpreted as a feature representation for the entire graph. $V=\{\mathbf{v}_i\}_{i = 1:n}$ is the set of vertices/nodes (cardinality $n$) where each $\mathbf{v}_i$ is a feature for node $i$, and $E = \{ ( \mathbf{e}_k, r_k, s_k ) \}_{k = 1:m}$ is the set of edge tuples (cardinality $m$) for which edge $k$ connects the source node with index $s_k$ to the target node with index $r_k$. For simplicity, we assume the global, vertex, and edge features have the same dimensionality $D$. That is, $\mathbf{u} \in \mathbb{R}^D$, $\mathbf{v}_i \in \mathbb{R}^D\ \, \forall \: i \in \{ 1, \ldots, n \}$, $\mathbf{e}_k \in \mathbb{R}^D\ \, \forall \: k \in \{ 1, \ldots, m \}$.

\begin{algorithm}[b!p]
\SetAlgoLined
\SetAlgoNoEnd
\SetKwFunction{FMain}{\textit{GraphNetwork}}
\SetKwProg{Fn}{function}{}{end function}
\SetKw{Let}{let}
{\small
\Fn{\FMain{$\mathbf{u}$, $V$, $E$}} {
\For{$k \in \{ 1 \ldots m \}$}{
\tcp{update edge features}
$\mathbf{e}'_k \leftarrow \phi^e(\mathbf{e}_k, \mathbf{v}_{r_k}, \mathbf{v}_{s_k}, \mathbf{u})$ 
}
\For{$i \in  \{ 1 \ldots n \}$}{
\tcp{aggregate incoming edges}
\Let $E'_i = \{ (\mathbf{e}'_k, r_k, s_k) \}_{r_k = i, k = 1:m}$ \\
$\mathbf{\bar{e}}'_i \leftarrow \rho^{e \rightarrow v}(E'_i)$ \\
$\mathbf{v}'_i \leftarrow \phi^v(\mathbf{\bar{e}}'_i, \mathbf{v}_i, \mathbf{u})$ \tcp{update node features}
}
\tcp{aggregate updated node/edge features}
\Let $V' = \{\mathbf{v}'_i \}_{i = 1:n}$ \\
$\mathbf{\bar{v}}' \leftarrow \rho^{v \rightarrow u}(V')$ \\
\Let $E' = \{ (\mathbf{e}'_k, r_k, s_k) \}_{k = 1:m}$ \\
$\mathbf{\bar{e}}' \leftarrow \rho^{e \rightarrow u}(E')$ \\
$\mathbf{u}' \leftarrow \phi^u(\mathbf{\bar{e}}', \mathbf{\bar{v}}', \mathbf{u})$ \tcp{update global feature}
\KwRet{($\mathbf{u}'$, $V'$, $E'$)}
}
}
\caption{Computation in a GN Block}
\label{alg:gn-block}
\end{algorithm}

The basic element of a GNN is a graph network block (GN block). A GN block takes as input a graph $\tilde{G} = (\tilde{\mathbf{u}}, \tilde{V}, \tilde{E})$ and produces an updated graph $\tilde{G}' = (\tilde{\mathbf{u}}', \tilde{V}', \tilde{E}')$ which can have arbitrary feature dimensionality. As detailed in Algorithm \ref{alg:gn-block}, the computation is done by first updating the edge features, followed by the node features and lastly the global features.
%
The update functions $\phi^v(\cdot)$, $\phi^e(\cdot)$, $\phi^u(\cdot)$ and aggregation functions $\rho^{e \rightarrow v}(\cdot)$, $\rho^{e \rightarrow u}(\cdot)$, $\rho^{v \rightarrow u}(\cdot)$ of the algorithm can be implemented in different ways. In particular, we use multi-layer perceptrons for implementing the update functions $\phi$. Since the aggregation functions must be symmetric and agnostic to the input permutation, we use the summation function, although averages or max/min could be used as well. We then compose our GNN using sequential GN blocks.

\subsection{Graph Representation}
\label{sec:graph_representation}

To use GNNs for localization, we must convert the concept of a topological map into the representation defined in Sec.~\ref{sec:gnn_overview}. We achieve this goal with learnable node, edge, and global features, represented as the embedding lookup table in Fig.~\ref{fig:graph-localizer}.
The feature for each node is one of three possibilities, depending on the node type: \textit{room}, \textit{hallway}, \textit{open space}. Similarly, each edge feature can be one of five options: \textit{corridor follow}, \textit{turn left}, \textit{turn right}, \textit{find door}, \textit{straight (into room)}. The global feature changes at each timestep and is a function of the current visual input -- see Sec.~\ref{sec:graph_localization_network} for more details. 
%
As the model is trained, the node features, edge features, and the CNN used for generating the global features are all learned jointly with the graph localization network.

\subsection{Graph Localization Network (GLN)}
\label{sec:graph_localization_network}

The goal of the graph localization network is to predict the robot's location in the map based on its current visual observation, its last predicted location, and the entire map represented as described in Sec.~\ref{sec:graph_representation}.
To accomplish this, we use a CNN to process the observations into visual features, which are used as the global feature in our graph representation. In parallel, we crop the graph to the local region around the last predicted location. Then, together with the node, edge, and newly computed global features, the graph is passed through the GNN to predict the agent's current edge in the graph. Note that when navigation begins, the agent is provided with its ground truth location (e.g. office 1). After the initial timestep, the agent relies on its own localization predictions.

\subsubsection{Computing the Global (Visual) Features}
\begin{figure*}[t]
    \centering
    \includegraphics[width=0.8\linewidth]{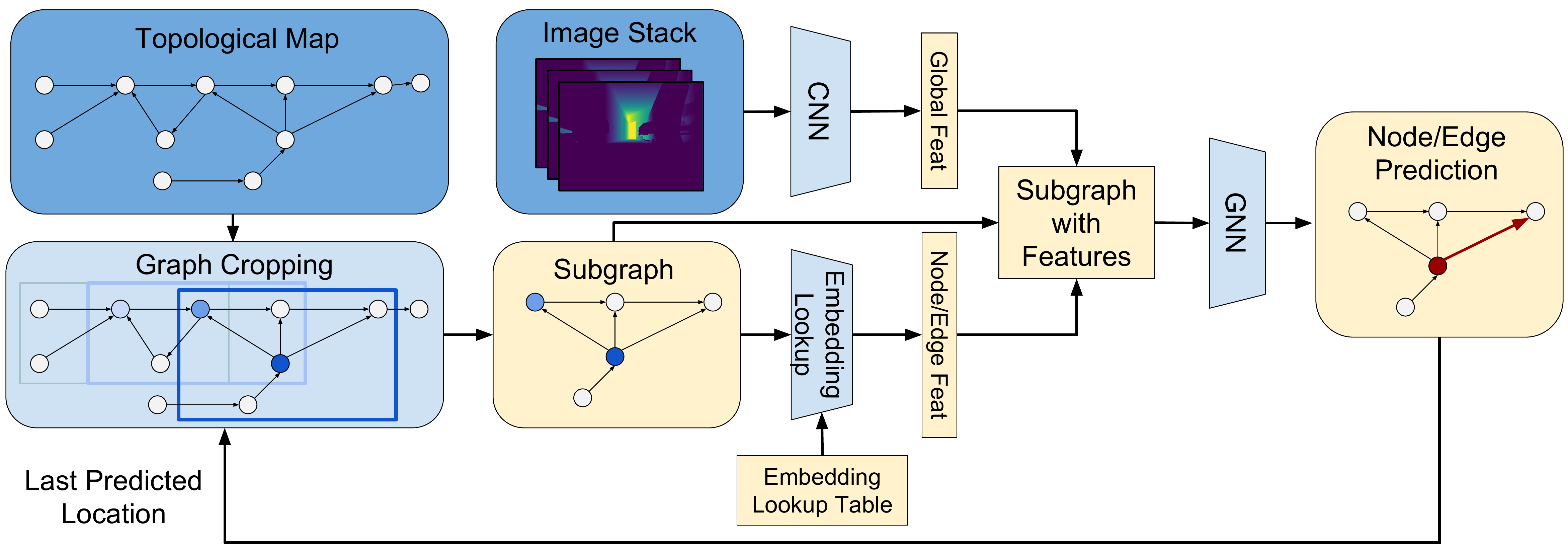}
    \caption{The graph localization network (GLN) takes three inputs: the depth image stack, the topological map, and the last predicted location. This information is then used to predict the agent's current position within the topological map.}
    \label{fig:graph-localizer}
    \vspace{-1em}
\end{figure*}

More concretely, the inputs to the graph localization network are the graph vertices and edges, the last predicted location, and an image stack $I$ of dimension $H \times W \times C$ where $H$ and $W$ are the image height and width, respectively. To ensure that spatio-temporal information is captured from the visual observations, the agent maintains a stack of the $C$ most recent depth image frames.\footnote{Although other modalities such as RGB may be used instead, we use depth images (clipped to a maximum distance of 3.5 m) to facilitate with generalization to different scenes which may have diverse visual/color appearances.} 
The image stack is forward passed through a convolutional neural network, as shown in Fig.~\ref{fig:graph-localizer}, to compute visual features. These features are used as the global feature $\mathbf{u} \in \mathbb{R}^D$ in our model's graph neural network.

\subsubsection{Subgraph Cropping}
In parallel to the computation of the global features, the topological map is mapped to its graph representation using the node and edge features described in Sec.~\ref{sec:graph_representation}. Since the graph of the entire environment can be very large and it is unlikely for the robot to move from one side of the graph to another far side, we crop a local region of the map centered on the previous predicted robot location. In particular, we crop a node if it is above a certain number of edges away from the previous predicted location. The localization prediction is then performed on the local subgraph, which also has the added benefit of reducing computation.

During training, the correct localization is always at the center of the subgraph, which may not necessarily hold true at test time due to noisy (previous) localization predictions. To increase localization robustness, we perform data augmentation during training by sampling nearby nodes as the center for the subgraph.


\subsubsection{Graph Neural Network Prediction}
To train the graph neural network, we treat the localization problem as a classification task. Given a subgraph $S$ with $m_s$ edges, the goal is to classify which of the $m_s$ edges the robot is currently on. In our setup, we use edge classification rather than node classification because edge classification is better defined. For example, at any instant the agent is executing an edge along the navigation plan. Additionally, the edge carries both source node and target node information, so it is trivial to localize the agent to a (source) node given an edge prediction.

Our network is composed of two sequential GN blocks, with the last block outputting per-edge logits of dimension 1. Let $m_s$ be the number of edges in the subgraph, $y$ be the index of the ground truth edge, and $\mathbf{p}^e$ be the vector of unnormalized probabilities such that each element $p^e_k$ is the unnormalized probability that the agent is on edge $k$. To train the network, we use a softmax cross-entropy loss on the edge probabilities: 
\[ l(\mathbf{p}^e, y) = - \log ( \exp{p^e_y} / \sum_k \exp{p^e_k} )\]


\subsection{Behavior Networks}
\label{sec:behaviornet}

Once the robot has been localized, the next question is how to control the robot given this coarse localization information. Unlike most prior deep-learning based approaches, we use a behavioral approach \cite{sepulveda2018deep} such that our action space is now composed of high-level semantic behaviors rather than low level motor control. 
By using a learning-based, data-driven approach, we can not only directly use the visual inputs for control but also circumvent the need for precise localization. This ties in nicely with the topological map representation because coarse localization allows planning a path in the topological map which directly translates to a sequence of behaviors to execute. The first action of this sequence determines which behavior network to use at the current timestep for low-level motor control. 


We implement the behavior networks as either a convolutional neural network or a recurrent neural network, depending on the specific behavior. The input to the network is the visual information (e.g. depth images) and the output is the control velocities for the robot. For the \textit{corridor follow} and \textit{find door} behavior, we use a CNN similar to the one used for computing the graph global features in Sec.~\ref{sec:graph_localization_network}. The input is an image stack $I$ with dimension $H \times W \times C$, and the output is the translational and rotational velocities $v = [v_p, v_\theta]$. For all other behaviors (\textit{turn right}, \textit{turn left}, \textit{straight}), we use a Long Short-Term Memory network \cite{hochreiter1997long} with a CNN encoder. These recurrent networks worked well in preliminary experiments. 
%
We collected a dataset (Sec.~\ref{sec:dataset-collection}) and trained these networks via behavioral cloning using a mean squared error loss on the predicted and ground truth velocities: $l(\nu, \hat{\nu}) = ( \nu - \hat{\nu} )^2$.

\begin{figure*}
    \centering
    \includegraphics[width=0.98\linewidth]{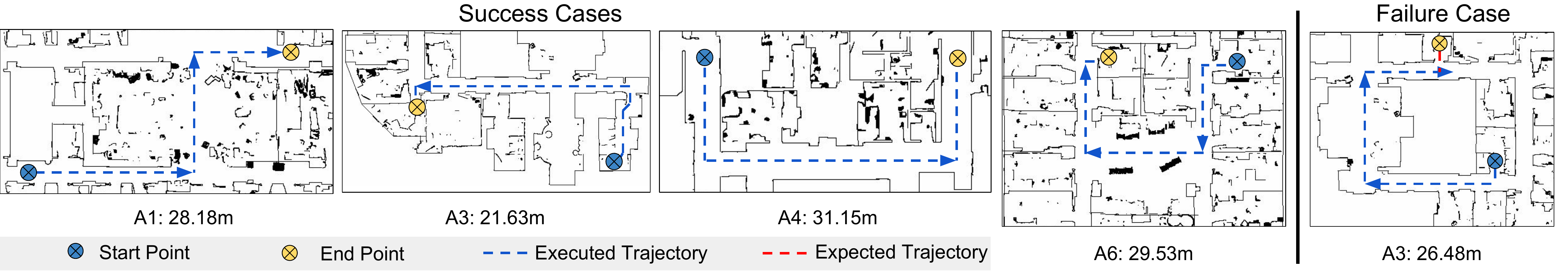}
    \caption{Examples of executed trajectories with the approximate navigation plan lengths. Seen environments: Areas 1, 5, 6 (A1, A5, A6). Unseen environments: Areas 3, 4 (A3, A4). Best viewed in color.}
    \label{fig:example-trajectories}
    \vspace{-1em}
\end{figure*}

\section{Experimental Setup}

We 
perform all training and testing in the Gibson simulator, which is powered by the Bullet physics engine \cite{xiazamirhe2018gibsonenv}. 
%
This setup is fairly different from those used by prior DL approaches for navigation. For example, several works use synthetically generated environments which do not accurately represent real indoor spaces \cite{mirowski2016learning,sepulveda2018deep,savinov2018semi}. Other approaches have been tested in more realistic house or office settings \cite{wu2018building,yang2018visual,gupta2017unifying,gupta2017cognitive}, but they ignore the collision problem entirely
and allow the agents to continue their trajectories despite undergoing collisions. 
In our case, collisions are fatal and result in failed navigation. 

We model the agent as a Turtlebot robot which is operated via ROS. The robot is equipped with a depth camera with a \ang{150} field-of-view. This wide angle view alleviates problems with occlusions and doorways. Commands are executed with a frequency of 5 Hz and the robot's velocity is capped at 0.5 m/s. We do not provide ground truth ego-motion to the agent, in contrast to prior works \cite{mirowski2016learning,gupta2017cognitive,parisotto2017neural}. 



\subsection{Dataset Collection}
\label{sec:dataset-collection}

We collected data within Gibson in order to train the behavior networks and graph localization network. In particular, we ran thousands of navigation tasks in simulation with the ROS Navigation Stack \cite{lu2014layered} using ground truth odometry and recorded visual observations from the robot (RGB, depth, and semantic information) as well as odometry information. We also injected noise into the velocity commands during data collection in order to teach the agent to recover from poor positioning. In our experiments, we used only depth images for practicality and generalizability to different environments, but provide RGB and semantic data for future endeavors.

After data collection, we used an automated annotation process based on heuristics, outlined in the supplementary material, to label the robot's trajectory data in relation to the environment's topological map. In particular, we labeled frames/timesteps with tags corresponding to the current behavior that is being executed by the robot, the current node/edge that it is traversing, and the semantic location (e.g., room name). 
In total, we collected 2,371 motion trajectories
with an average of 423.56 frames per trajectory.

\subsection{Navigation Evaluation Suite}
\label{sec:eval-suite}

Because our automated process to label robot trajectories can run in real-time, we were able to create an experimental setup for the systematic evaluation of behavioral navigation approaches. Our setup aims to facilitate reproducibility, such that we can easily perform a thorough analysis of the performance of various navigation models in realistic indoor environments. 
For example, our setup can identify that a navigation approach is very effective at turning from a hallway into another hallway, but struggles with turning from a hallway into an office.
We refer to this infrastructure 
as our \textit{evaluation suite}, and further detail its key features in the next paragraphs.


\subsubsection{Generation of Navigation Tasks}
Our evaluation suite supports sampling navigation plans (random start and end nodes, followed by the shortest path) 
for the 
Stanford 2D-3D-S dataset for which we created topological maps. Sampled plans can then be used to evaluate navigation approaches or roll-outs of navigation policies.
In our experiments, we generated a set of navigation tasks separately from the previously mentioned dataset of Sec. \ref{sec:dataset-collection}. Their path lengths range from very short ($< 5$ meters) to very long ($> 50$ meters).  Example trajectories resulting from these tasks can be observed in Fig.~\ref{fig:example-trajectories}. 

\subsubsection{Evaluation Metrics} 
Our evaluation suite supports several kinds of metrics. First, it can evaluate navigation performance based on \textit{success rate}, similar to prior work \cite{sepulveda2018deep,savva2017minos}. A success occurs if the robot can follow the navigation plan all the way until the destination without deviating from the path. Conversely, a failure occurs if the robot does deviate or gets stuck due to a collision. Second, our suite can evaluate partial plan completion. More specifically, \textit{plan completion} is measured as the fraction of nodes in the plan that were successfully reached by the agent during navigation. Third, performance evaluation can be conducted in 
semantically meaningful ways. 
For example, 
our suite allows to check the average success rate of a particular \textit{turn} or \textit{junction} in the plan or map, or whether the robot struggles more with turning \textit{into} offices compared to turning \textit{out of} them. These metrics are particularly useful as they provide insight into 
what kinds of scenarios an approach is likely to succeed or fail in.

\subsection{Implementation Details}

We implemented all neural networks using PyTorch \cite{paszke2017automatic}. 
We use depth images of $320 \times 240$ pixels and image stacks of the $C = 20$ most recent frames where appropriate. Each behavior network is trained individually.  
For training all networks, we use the Adam optimizer~\cite{kingma2014adam} with a learning rate and batch size of 1e-4 and 32, respectively.  We implement the CNNs as a series of strided convolution layers with batch normalization \cite{ioffe2015batch}, and the graphs are encoded using global, node, and edge features of dimension 512. The supplementary material provides more details. 

\section{Experimental Results}

We conduct experiments to evaluate the performance of 
our navigation approach against several baselines.
Following \citet{gupta2017unifying}, we use five areas \{1, 3, 4, 5, 6\} from the Stanford 2D-3D-S dataset in our evaluation. Areas 1, 5, and 6 are used for training, area 3 is used for validation, and area 4 is used for testing. It is worth noting that there are only three distinct buildings in the dataset: areas 1, 3, 6 correspond to different parts of one building, and area 4 and area 5 are each captured in different buildings. Because each area is unique, with varying size and structure, we report results for all of the train, validation, and test areas. 

We divide the navigation tasks into three difficulties based on the number of nodes in the corresponding path: 1 through 10 nodes corresponds to difficulty I; 11 through 20 nodes is difficulty II; and $>20$ nodes is difficulty III (see the supplementary material for additional details). 

Our evaluation considers 3 baselines:
\begin{description}[font=\textbf,style=unboxed,leftmargin=0cm,noitemsep]
\item[PhaseNet:] This network \cite{yu2018one} determines when to transition behaviors by predicting their \textit{phase}, or temporal progress. We implement this network with an LSTM trained with a mean-squared error objective on the progress of the behavior.
\item[BehavRNN:] Sequence-to-sequence deep learning model \cite{sutskever2014sequence} trained to perform behavior classification at each timestep with a softmax cross-entropy loss. The model takes as input the current visual observation and the navigation plan (as a sequence of behaviors).
\item[GTL:] Navigation approach that uses our automated annotation tools for Gibson to compute the Ground Truth Location (GTL) of the robot in real-time relative to the  map. To navigate, the robot executes the behavior network according to its current position in the map. The behavior networks used for this baseline are the same as in our approach.
\end{description}
The first two baselines serve to compare our approach with other relevant deep learning methods for behavioral navigation. The third baseline helps us study the performance of our behavior networks in isolation from potential localization errors induced by our graph localization network. We refer to our approach as \textit{GraphNav} in our experiments. In addition, we implemented a particle filter for filtering the GLN predictions and refer to this model as \textit{GraphNavPF}.

We evaluate approaches based on full plan success rates, per-behavior success rates, and average plan completion. As mentioned before, plan completion is computed as the fraction of nodes in the plan that the agent successfully reached. 
\begin{table*}[t]
    \centering
    \setlength\tabcolsep{5pt}
    \begin{tabular}{ccccccccccc}
        \toprule
         & & \multicolumn{5}{c}{Per-Behavior Success Rates} & \multicolumn{3}{c}{Per-Difficulty SR / PC} & Total\\
        \cmidrule(l){3-7} \cmidrule(l){8-10}
        Area ID & Model & cf & fd & tr & tl & s & I & II & III & SR / PC \\
        \midrule
        1 (Seen) & PhaseNet & 89.5 (196) & 96.3 (54) & 39.0 (41) & 42.9 (42) & 0 (2) & 16.7 / 52.2 & 2.4 / 37.0 & 0 / 25.7 & 7.3 / 41.2 \\
        5 (Seen) & PhaseNet & 86.7 (158) & \textbf{93.3} (45) & 63.9 (36) & 48.1 (52) & - (0) & 9.1 / 54.0 & 0 / 37.3 & 0 / 21.4 & 3.0 / 40.9 \\
        6 (Seen) & PhaseNet & 87.3 (173) & \textbf{98.1} (53) & 34.2 (38) & 36.8 (57) & 100 (2) & 10.3 / 52.5 & 2.4 / 37.6 & 0 / 18.5 & 5.6 / 39.9 \\
        1 (Seen) & BehavRNN & 72.4 (127) & 94.4 (54) & 43.8 (32) & 31.3 (32) & \textbf{50.0} (2) & 10.0 / 48.9 & 0 / 23.3 & 0 / 12.7 & 3.7 / 31.4 \\
        5 (Seen) & BehavRNN & 73.8 (107) & 82.2 (45) & 58.8 (34) & 55.6 (36) & - (0) & 0 / 40.4 & 0 / 27.7 & 0 / 17.2 & 0 / 30.6 \\
        6 (Seen) & BehavRNN & 69.3 (137) & 92.3 (52) & 56.3 (32) & 51.1 (52) & 0 (2) & 12.8 / 55.0 & 0 / 28.4 & 0 / 13.9 & 5.6 / 36.3 \\
        1 (Seen) & GraphNav (ours) & 91.4 (441) & \textbf{98.1} (54) & \textbf{98.2} (55) & \textbf{79.0} (62) & 20.0 (5) & 53.3 / 81.9 & 19.0 / \textbf{60.9} & \textbf{10.0} / \textbf{58.0} & 30.5 / \textbf{68.2} \\
        5 (Seen) & GraphNav (ours) & 97.7 (344) & 91.1 (45) & 70.8 (65) & 76.6 (64) & \textbf{100} (1) & 36.4 / 77.1 & 30.6 / 60.9 & 12.5 / 66.0 & 30.3 / 66.9 \\
        6 (Seen) & GraphNav (ours) & 94.3 (388) & 96.2 (53) & 74.6 (67) & 83.1 (77) & 50.0 (6) & 51.3 / 80.8 & 22.0 / 63.1 & \textbf{33.3} / \textbf{59.6} & 36.0 / 68.3 \\
        1 (Seen) & GraphNavPF (ours) & \textbf{91.7} (409) & \textbf{98.1} (54) & 94.4 (54) & 75.9 (58) & 20.0 (5) & \textbf{56.7} / \textbf{82.3} & \textbf{21.4} / 58.9 & 0 / 48.1 & \textbf{31.7} / 66.2 \\
        5 (Seen) & GraphNavPF (ours) & \textbf{98.6} (420) & \textbf{93.3} (45) & \textbf{90.3} (72) & \textbf{86.7} (75) & \textbf{100} (3) & \textbf{59.1} / \textbf{82.8} & \textbf{66.7} / \textbf{79.1} & \textbf{37.5} / \textbf{69.8} & \textbf{60.6} / \textbf{79.2} \\
        6 (Seen) & GraphNavPF (ours) & \textbf{94.8} (430) & 96.2 (52) & \textbf{87.8} (74) & \textbf{94.0} (84) & \textbf{62.5} (8) & \textbf{69.2} / \textbf{84.8} & \textbf{43.9} / \textbf{77.4} & \textbf{33.3} / 53.4 & \textbf{53.9} / \textbf{75.9} \\
        1 (Seen) & GTL\textsuperscript{\textdagger} & 91.3 (438) & 98.2 (54) & 92.9 (56) & 88.5 (61) & 25.0 (4) & 56.7 / 81.0 & 23.8 / 64.1 & 20.0 / 53.1 & 35.4 / 68.9 \\
        5 (Seen) & GTL\textsuperscript{\textdagger} & 99.8 (523) & 93.3 (45) & 95.1 (82) & 96.6 (88) & 100 (4) & 68.2 / 90.0 & 91.7 / 97.0 & 87.5 / 88.1 & 83.3 / 93.6 \\
        6 (Seen) & GTL\textsuperscript{\textdagger} & 94.3 (418) & 98.2 (54) & 83.8 (74) & 97.7 (85) & 55.6 (9) & 64.1 / 84.7 & 46.3 / 77.5 & 22.2 / 49.8 & 51.7 / 75.6 \\
        \midrule
        3 (Unseen) & PhaseNet & 87.6 (121) & \textbf{100} (33) & 40.0 (40) & 46.3 (41) & \textbf{100} (2) & 20.4 / 61.3 & 0 / 37.8 & - / - & 15.3 / 55.4 \\
        3 (Unseen) & BehavRNN & 69.2 (117) & 97.0 (33) & 62.9 (35) & 63.2 (38) & \textbf{100} (1) & 14.8 / 54.9 & 0 / 40.0 & - / - & 11.1 / 51.1 \\
        3 (Unseen) & GraphNav (ours) & 92.3 (182) & 97.0 (33) & 57.4 (47) & 76.5 (51) & \textbf{100} (3) & 40.7 / 75.5 & 16.7 / 61.2 & - / - & 34.7 / 71.9 \\
        3 (Unseen) & GraphNavPF (ours) & \textbf{95.6} (206) & \textbf{100} (33) & \textbf{70.0} (50) & \textbf{78.0} (59) & 75.0 (4) & \textbf{50.0} / \textbf{77.6} & \textbf{38.9} / \textbf{78.1} & - / - & \textbf{47.2} / \textbf{77.7} \\
        3 (Unseen) & GTL\textsuperscript{\textdagger} & 96.1 (228) & 97.0 (33) & 98.2 (54) & 92.1 (63) & 75.0 (4) & 74.1 / 86.6 & 83.3 / 88.9 & - / - & 76.4 / 87.2 \\
        \midrule
        4 (Unseen) & PhaseNet & 90.1 (192) & 81.3 (32) & 45.8 (48) & 61.1 (36) & 0 (1) & 17.9 / 55.4 & 9.3 / 37.2 & 0 / \textbf{24.5} & 12.0 / 41.9 \\
        4 (Unseen) & BehavRNN & 76.4 (140) & 87.5 (32) & 43.6 (39) & 57.7 (26) & - (0) & 14.3 / 49.5 & 2.3 / 27.1 & 0 / 9.3 & 6.7 / 34.5 \\
        4 (Unseen) & GraphNav (ours) & \textbf{90.5} (252) & 90.6 (32) & 55.4 (56) & \textbf{76.1} (46) & - (0) & 25.0 / 67.8 & \textbf{11.6} / \textbf{44.8} & 0 / 17.7 & 16.0 / \textbf{52.0} \\
        4 (Unseen) & GraphNavPF (ours) & 87.5 (232) & \textbf{93.75} (32) & \textbf{63.6} (55) & 75.0 (44) & - (0) & \textbf{32.1} / \textbf{68.7} & 9.3 / 41.7 & 0 / \textbf{24.5} & \textbf{17.3} / 50.9 \\
        4 (Unseen) & GTL\textsuperscript{\textdagger} & 95.7 (376) & 87.5 (32) & 81.1 (74) & 92.5 (67) & - (0) & 57.1 / 77.6 & 46.5 / 72.0 & 0 / 33.0 & 48.0 / 72.0 \\
        \bottomrule
    \end{tabular}
    \caption{Performance comparison using success rate (SR) and average plan completion (PC). Numbers in parentheses represent total number of attempts for that entry. GraphNavPF uses a particle filter on the GLN predictions (see the supplementary material). The \textsuperscript{\textdagger} indicates that GTL utilizes additional ground truth information. The dashes (-) indicate that there were no trajectories for the corresponding entry, and best performing entries are bolded per area (not including GTL).}
    \label{tab:quantitative-results}
    \vspace{-1em}
\end{table*}

\begin{figure*}
    \centering
    \includegraphics[width=0.9\linewidth]{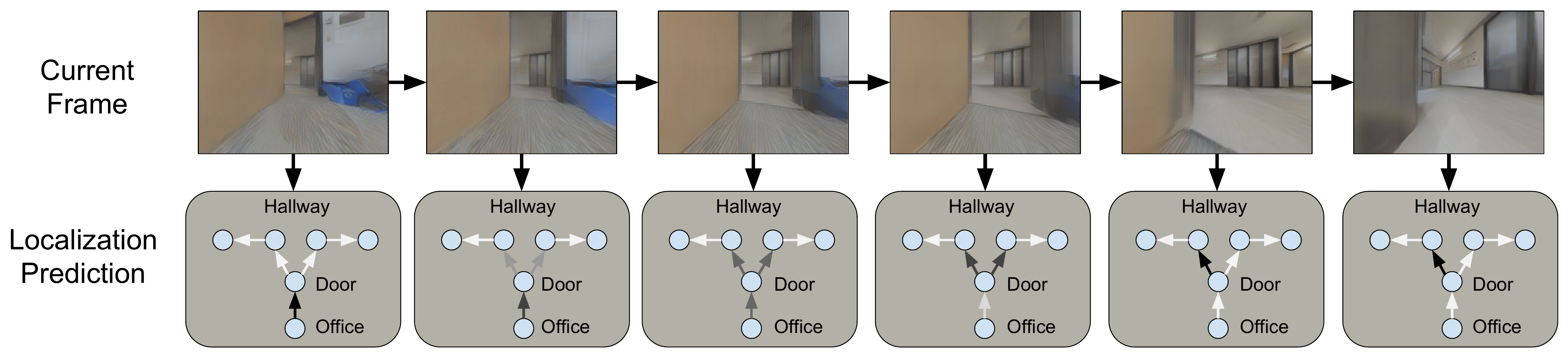}
    \caption{GLN predictions over time. Darker arrows indicate higher probability predictions. Initially, the network localizes the agent to the office. As the robot approaches the door, the predictions change accordingly. Once the robot reaches the door, the controller executes the turn left behavior, affecting the visual information and changing the localization prediction in turn.
    }
    \label{fig:localization-visual}
    \vspace{-1.5em}
\end{figure*}

\subsection{Overall Navigation Performance}



The quantitative results can be found in Table~\ref{tab:quantitative-results}. The results show that  PhaseNet works very poorly, resulting in the lowest success rates and plan completion percentages. While the \textit{corridor follow} and \textit{find door} behaviors have fairly high per-behavior success rates, the turning success rates are very low, usually below 50\%. This approach results in an agent that tends to follow corridors blindly, missing most of the turns.

Our approach outperforms the PhaseNet and BehavRNN baselines in all areas. Relative to the PhaseNet baseline, the turn success rates are significantly better in both seen and unseen areas, such as 42.9\% (PhaseNet) vs. 79.0\% (GraphNav) for turning left in area 1.
Similarly, the success rates and plan completion percentages across difficulty levels are higher for our GraphNav model. While this difference is not as big in the unseen areas, there is still a substantial difference in performance. For example, PhaseNet achieves an average plan completion of 55.4\% in unseen area 3, whereas GraphNavPF achieves an average of 77.7\%.

Fig.~\ref{fig:localization-visual} shows a qualitative example of how the localization works with GraphNav. While the agent is in the office, the graph localization network correctly predicts the location. As the robot approaches the door, it is unclear in which direction it will turn. 
At this point, the network weighs the left and right turns equally, which translates to predicting that the agent is at the door node, triggering a behavior transition. Lastly, as the visuals show the robot turning left, the GLN becomes more confident that it is on the edge corresponding to the left turn. 

Successful navigation tasks by GraphNav are shown in the four left-most images of Fig.~\ref{fig:example-trajectories}. The robot completes trajectories ranging from 20 m to 32 m. On the right is a failure case in which the robot navigated most of the 26 meter-long  trajectory but deviated from the path towards the end. 


\subsection{Performance of the Behavior Networks}
\label{sec:behaviornet-performance}

Qualitatively, we observe that the behavior networks used in our approach succeed in their assigned task (e.g., follow a corridor, turn left, turn right) while being robust to collisions with walls and clutter, especially in structured areas. 
We refer the reader to our webpage (\url{https://graphnav.stanford.edu}) for more qualitative examples.

We verify our observations quantitatively using the GTL baseline, which uses our behavior networks for motion control along with our annotation tool for localization.
As can be seen in Table~\ref{tab:quantitative-results}, GTL performs well, especially in structured environments such as area 5 (train), which has an average plan completion of 93.6\% and area 3 (val), which has an average plan completion of 87.2\%. Examining the per-behavior success rates, we observe results generally above 80\% and even 90\%, indicating robustness in both seen and unseen environments.

GTL 
struggles in certain cases. One example is large open spaces, 
which are prevalent in areas 1, 3, 4, and 6. In these spaces, the robot sometimes fails to orient itself correctly during behavior execution, and walks into a corner 
or deviates from the correct path. 
Recovery is difficult because the clutter prevents the agent from having a direct line-of-sight to the room exit. This challenge may be due to the maximum depth of the robot's observations (3.5 m).

\section{Conclusion and Future Work} 
\label{sec:conclusion}

We introduced an effective topological map design for behavioral navigation and, to the best of our knowledge, are the first to propose graph neural networks for robot localization. We tested our proposed approach using Gibson, and provide an open-source testbed for benchmarking navigation in complex, human environments. Our results show the potential of combining DL with classical robotic architectures.

While our method was able to outperform other DL baselines, there are several, interesting future research directions. First, our topological maps were manually annotated and the behaviors were pre-defined, limiting the scalability of our setup.
In the future, it would be interesting to investigate mechanisms to create topological maps and behaviors in a data-driven fashion.
%
%
Second, there is room for improvement in terms of navigation success rate (Table~\ref{tab:quantitative-results}). 
One key challenge that is worth further investigation is the timing of the transitions between behaviors. The use of more explicit semantic information could also be advantageous in our problem setting. 
Lastly, future work could investigate sim-to-real transfer and conduct experiments in real environments with a more practical camera setup. We hope that our work inspires further research to advance autonomous, visual navigation.


\section{Acknowledgements} 
\label{sec:ack}

Toyota Research Institute (``TRI'')  provided funds to assist the authors with their research but this article solely reflects the opinions and conclusions of its authors and not TRI or any other Toyota entity. This work is also partially funded by Fondecyt grant 1181739, Conicyt, Chile.

\bibliographystyle{plainnat}

\newpage


\begin{appendices}
    \section{Implementation Details}

We used the dataset described in Sec. VI-A of the main paper to train the behavior networks and the graph localization network individually (Fig. 3 of main paper). The networks were all trained using the Adam optimizer \cite{kingma2014adam} with a learning rate of 1e-4 and a batch size of 32, as mentioned in Sec. VI-C of the main paper.

For all convolutional layers, we used convolutions without padding, and we used ReLU for all activation functions. All networks were implemented using PyTorch \cite{paszke2017automatic}.

\subsection{Graph Localization Network (GLN)}

\subsubsection{Subgraph Cropping}
During training time, we crop the graph by first sampling a node to be the center for the crop region, as discussed in Sec. V-C of the main paper. We then remove all nodes from the graph that are beyond a distance of 3 edges ahead of the sampled node or a distance of 2 edges behind sampled node. Lastly, we perform a check to ensure the ground truth edge is within the subgraph. If the subgraph does not contain the ground truth edge, we re-sample a new crop center and repeat the process until a valid subgraph is obtained. At test time, we use the source node from the previous edge prediction as the crop center.

\subsubsection{Computing the Global (Visual) Features}
\label{sec:global-features}

A convolutional neural network (CNN) processes the input depth image stack of size $320 \times 240 \times 20$, where the channels represent the 20 most recent depth images, captured at a rate of 5 Hz. The architecture of the CNN is described in Table~\ref{tab:gln-cnn}. The output of the network is a feature set of shape $1 \times 1 \times 512$, which can simply be reshaped into a 512-dimnension feature vector that serves as a global feature for the graph neural network (Sec.~\ref{sec:gnn}).

\begin{table}[b]
    \centering
    \begin{tabular}{ccccc}
        \toprule
        Layer & Channels Out & Kernel & Stride & BN \\
        \midrule
        conv1 & 32 & $5 \times 5$ & 2 & Yes \\
        conv2 & 64 & $5 \times 5$ & 2 & Yes \\
        conv3 & 128 & $3 \times 3$ & 2 & Yes \\
        conv4 & 256 & $3 \times 3$ & 2 & Yes \\
        conv5 & 512 & $3 \times 3$ & 2 & Yes \\
        conv6 & 512 & $3 \times 3$ & 2 & Yes \\
        conv7 & 512 & $3 \times 2$ & 1 & No \\
        \bottomrule
    \end{tabular}
    \caption{Architecture of the CNN encoder used for the GLN global (visual) feature computation and for the CNN-LSTM behavior network.}
    \label{tab:gln-cnn}
\end{table}

\subsubsection{Graph Neural Network}
\label{sec:gnn}
The graph neural network is composed of two sequential GN blocks and takes the node, edge, and global features as input. The global features are computed using the CNN described in Sec.~\ref{sec:global-features}. The update functions $\phi^v(\cdot)$, $\phi^e(\cdot)$, $\phi^u(\cdot)$ described in the main paper are implemented as multi-layer perceptrons (MLPs), outlined in Table~\ref{tab:gln-update-fn}. The aggregation functions $\rho^{e \rightarrow v}(\cdot)$, $\rho^{e \rightarrow u}(\cdot)$, $\rho^{v \rightarrow u}(\cdot)$ are elementwise sum functions.

\begin{table}[b]
    \centering
    \begin{tabular}{cccc}
        \toprule
        Layer & Output dim & BN \\
        \midrule
        fc1 & 256 & Yes \\
        fc2 & 256 & Yes \\
        fc3 & 256 & Yes \\
        fc4 & 256 & No \\
        \bottomrule
    \end{tabular}
    \caption{Architecture for the edge, node, and global feature update functions.}
    \label{tab:gln-update-fn}
\end{table}

\subsection{Behavior Network}

\begin{table}[b]
    \centering
    \begin{tabular}{ccccc}
        \toprule
        Layer & Channels Out & Kernel & Stride & BN \\
        \midrule
        conv1 & 32 & $5 \times 5$ & 2 & Yes \\
        conv2 & 64 & $5 \times 5$ & 2 & Yes \\
        conv3 & 128 & $3 \times 3$ & 2 & Yes \\
        conv4 & 256 & $3 \times 3$ & 2 & Yes \\
        conv5 & 512 & $3 \times 3$ & 2 & Yes \\
        conv6 & 512 & $3 \times 3$ & 2 & Yes \\
        conv7 & 2 & $3 \times 2$ & 1 & No \\
        \bottomrule
    \end{tabular}
    \caption{Architecture of the CNN behavior network.}
    \label{tab:cnn-behavior-network}
\end{table}

\begin{figure}
    \centering
    \includegraphics[width=\linewidth]{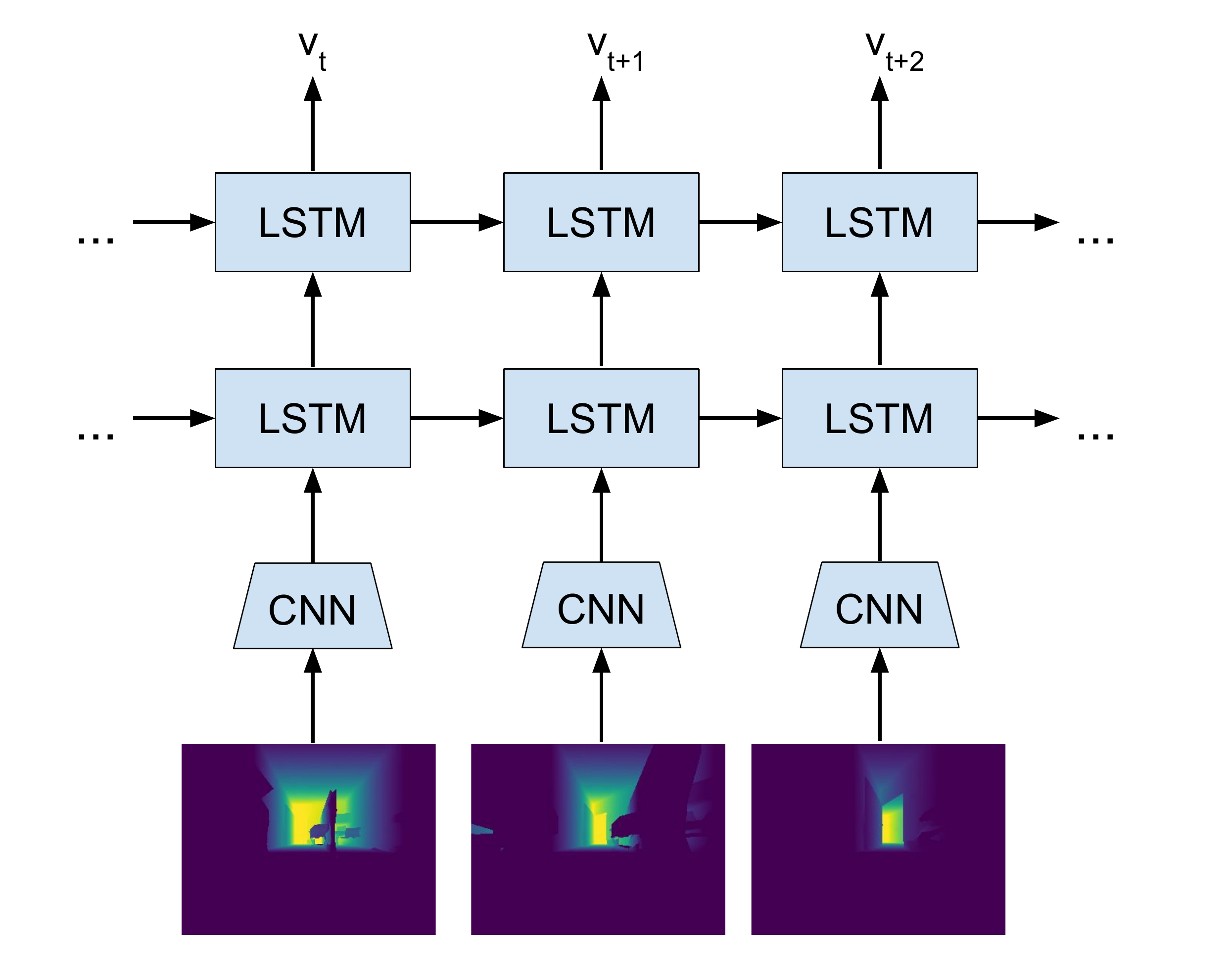}
    \caption{An illustration of our LSTM-based behavior network. The input at each timestep is the current depth image, and the output is the control velocities for the robot.}
    \label{fig:behaviornet-diagram}
\end{figure}

To predict velocity commands given a visual input, we use two types of behavior networks in our experiments: a purely reactive CNN, and a CNN-LSTM. The architecture of the CNN behavior network is shown in Table~\ref{tab:cnn-behavior-network}, and is used for the \textit{corridor follow} and \textit{find door} behaviors. The CNN-LSTM is used for the \textit{turn left}, \textit{turn right}, \textit{straight} behaviors, and has the structure shown in Fig.~\ref{fig:behaviornet-diagram}. A CNN is used to encode a single $320 \times 240 \times 1$ depth image, and the features are passed onto a two-layer long-short term memory (LSTM) \cite{hochreiter1997long} with a hidden size of 512. Lastly, the output of the LSTM goes through a fully-connected layer that outputs the translational and angular velocities: $v = [v_p, v_\theta]$.

\subsection{Particle Filter for GLN}

To improve the performance of GraphNav, we can combine our Graph Localization Network (GLN) with a particle filter and obtain more accurate localization predictions in a manner similar to \citet{dellaert1999monte} and \citet{fox1999monte}. In particular, the particle filter relies on distributions $p(x_t | u_t, x_{t-1})$ and $p(z_t | x_t)$, where $x_t$ represents the state vector (e.g. pose or location) at time $t$, $u_t$ is a known control input, and $z_t$ is a measurement at time $t$.

In particular, we represent the state $x_t$ as the current node in the topological map, and $z_t$ is the current visual observation. For any two consecutive timesteps $t$ and $t+1$, the executed control input (e.g. behavior) has little to no effect on the \textit{topological} location. Thus, we simplify the model and make the following assumption about the motion model:
\begin{equation}
    p(x_t | u_t, x_{t-1}) = p(x_t | x_{t-1})
\end{equation}
In our experiments, we found that a probability distribution with $p(x_t = x_{t-1} | x_{t-1}) = 0.8$ worked well (with equally weighted probabilities on the neighbors of $x_{t-1}$ and 0 for all other nodes).

For the measurement model $p(z_t | x_t)$, we make the assumption that $p(z_t)$ and $p(x_t)$ are uniform distributions for all timesteps, so the ratio $p(z_t) / p(x_t)$ can be written as a constant $\gamma$. Using Bayes rule, we obtain the following:
\begin{align*}
    p(z_t | x_t) &= \frac{p(x_t | z_t) p(z_t)}{p(x_t)} \\
    &=  \gamma p(x_t | z_t) \\
    &\propto p(x_t | z_t)
\end{align*}
We approximate $p(x_t | z_t)$ using the graph localization network (GLN) by aggregating (sum) the outgoing edge probabilities for each node. We use the motion and measurement models defined above and denote the GraphNav model with the particle filter as \textit{GraphNavPF}.

\section{Data Collection and Automated Dataset Annotation}

\begin{figure}
    \centering
    \includegraphics[width=\linewidth]{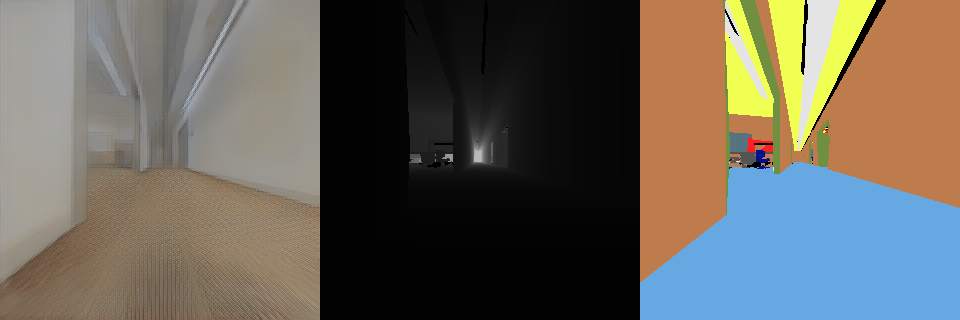}
    \caption{Left to right: Example RGB, depth, and semantics observations recorded with Gibson \cite{xiazamirhe2018gibsonenv}. Note that in our approach we only use the depth information.}
    \label{fig:dataset-collection-visual-input}
\end{figure}

Using the Gibson simulator \cite{xiazamirhe2018gibsonenv}, we sampled motion trajectories with the Turtlebot robot in the Stanford 2D-3D-S dataset \cite{armeni_cvpr16,armeni_arxiv_2d3ds}. Data collection was achieved as follows: we provided the ROS navigation stack with ground truth localization and used it to follow navigation plans, recording the robot's visual observations and ground truth odometry as it moved in the environment. An example of the collected visual input is shown in Fig.~\ref{fig:dataset-collection-visual-input}. After collecting the dataset, which contains 2,371 trajectories, we used an automated heuristic-based annotation algorithm to label frames with behavior ID tags and node/edge localizations. There are a few parameters that were used for automated labeling, and these were determined empirically. The datasets, annotations, and code described in this section will be presented to the public.



\subsection{Velocity Noise Injection}

We recorded the velocity commands from the ROS navigation stack as ground truth, but injected noise into the actual executed velocity commands provided to the robot. This was done in order to augment the dataset with more challenging scenarios that could be seen at test time. By using this method of data augmentation, we increase the diversity of scenarios the model sees at train time in the hope that this distribution matches more closely with the test time distribution. This method of data augmentation is similar to that used by \citet{bojarski2016end}.

To perform the noise injection, we use the following formula:
\begin{equation}
    \nu_{noisy} = \nu_{raw} + z
\end{equation}
Here, $\nu_{raw}$ represents the velocity commands from the ROS navigation stack (which are used for training), and $\nu_{noisy}$ are the noisy velocity commands used to actually control the robot during data collection. To prevent drastic changes in the movement of the robot, we update $z = [z_p, z_\theta]$ at every timestep using:
\begin{equation}
    z \leftarrow 0.95z + 0.05n
\end{equation}
where $n = [n_p, n_\theta]$ is sampled using $n_p \sim \mathcal{N}(0, 0.2)$ and $n_\theta \sim \mathcal{N}(0, 1)$.

\subsection{Room Annotations}

\begin{figure}
    \centering
    \includegraphics[width=0.7\linewidth]{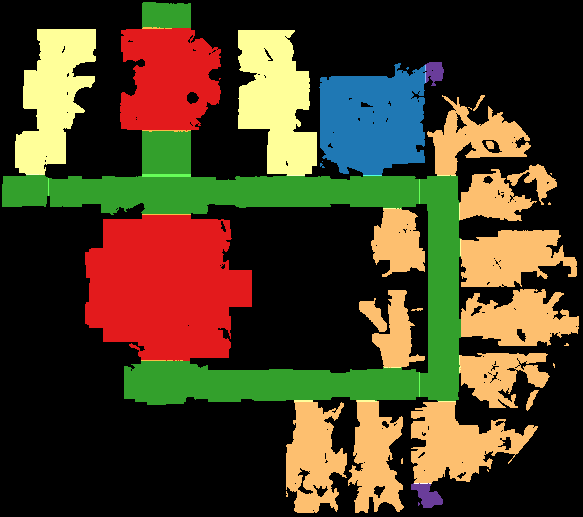}
    \caption{Birds-eye visualization of the Area 3 floorplan where different colors represent different semantic labels (hallway, bathroom, office, conference room, etc.).}
    \label{fig:area-3-semantic-floorplan}
\end{figure}

Using labels from the Stanford 2D-3D-S dataset, we were able to generate metric maps in the form of 2D images showing the clutter, semantic room labels, and instance-level room labels. An example of the semantic room labels is shown in Fig.~\ref{fig:area-3-semantic-floorplan}. With these metric maps, we were able to look up the instance-level room label for the robot given a position in the map. This was essential for the annotation and evaluation process. For example, the first step in processing our dataset for labels was to use the ground truth robot odometry to look up the instance-level room labels. These labels were then used for detecting behaviors, as explained in Sec.~\ref{sec:behavior-detection}.

\subsection{Behavior Detection}
\label{sec:behavior-detection}

To train our behavior networks, we required annotated video sequences of the agent executing each behavior. After annotation, each behavior network was trained using only the annotated frames for that particular behavior. For example, the \textit{corridor follow} behavior was trained using a collection of video clips from our dataset which contain the corridor follow behavior. Throughout this section, unless otherwise indicated, we refer to a room as any labeled instance-level region within the Stanford 2D-3D-S environments, such as a hallway, open space, office, or storage room.

We first processed the dataset to look for turn and straight behaviors. To do this, we found transition points for when the agent switched from one room to another. Given a transition point, we then checked whether it was a turn behavior by examining the agent's trajectory and checking if the robot rotated by greater than 40 degrees before traveling a distance of 2 m from the transition point. If it had rotated by over 40\degree, then this transition point was considered a turn in the corresponding direction. For each transition point, we performed this check twice -- one in forward time and one in backwards time from the transition point. The check with the larger rotation delta, compared to the robot's orientation at the transition point, was used for determining which behavior occurred at that particular transition point.

If no turn was detected for the transition point, we then checked whether the agent entered an actual room (office, conference room, pantry, bathroom, copy room, or storage room). If so, the transition point was considered a \textit{straight} behavior. Otherwise, it was considered a \textit{corridor follow} behavior. All frames within 2 m of the agent position at the transition point were then tagged with the behavior ID, except frames in which the robot entered a separate room (that was different from the transition rooms).

Next, we labeled frames with the find door behavior. Since all of the frames were labeled with a room name from the Stanford 2D-3D-S dataset, we could detect the first transition from a room to a hallway for each trajectory. In this particular case, rooms refer to offices, conference rooms, copy rooms, storage rooms, pantries, and bathrooms. All frames starting from the first frame of the trajectory to the transition frame were labeled as the \textit{find door} behavior.

The corridor follow behavior detection was run last. All untagged frames that were located in openspaces, hallways, and lounges were labeled to be the \textit{corridor follow} behavior. These labeled behavior frames were then used for training the behavior networks.

\subsection{Node and Edge Localization}

\begin{figure}[b]
    \centering
    \includegraphics[width=0.9\linewidth]{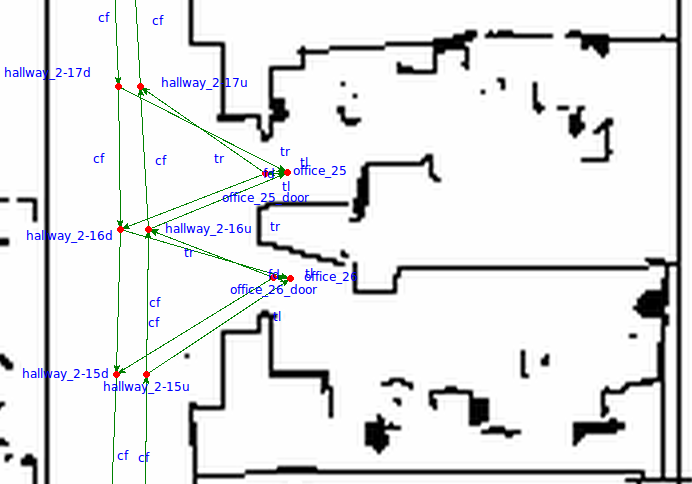}
    \caption{Topological map annotations using our graph drawer tool.}
    \label{fig:sem-graph-drawer}
\end{figure}

As mentioned throughout the main paper, we annotated the Stanford 2D-3D-S areas with topological map annotations in the form of a graph. These node and edge annotations were used for training the graph localization network, for running the GTL model, and for comparing our model with the baselines. To create the graph annotations, we wrote a graph drawer GUI tool that allowed us to draw nodes and edges on top of the metric map of each Stanford 2D-3D-S area (Fig.~\ref{fig:sem-graph-drawer}). These graph annotations contain node positions and edge orientations in the coordinate frame of the metric map (e.g. Fig.~\ref{fig:area-3-semantic-floorplan}). The positions and orientations were then used for annotating the dataset trajectories with corresponding nodes and edges in the graph.

For a given robot pose (position and orientation), we first computed the associated node and then the associated edge. At a high level, the node label is computed by finding the closest node (measured with euclidean distance) to the robot that has a similar orientation. Moreover, throughout each trajectory, we make sure that the node for any given frame is either a direct neighbor of the node in the previous frame or the same node as that of the previous frame.

More specifically, the first step to computing the associated node was to find the nearest node that satisfied a matching-orientation requirement and a same-room requirement. A node satisfies the matching orientation requirement if any of its incoming or outgoing edges is within 36\degree of the robot's orientation. The same-room requirement is satisfied if the node is located in the same room as the agent when the agent is in a room. If these requirements were satisfied, then we would move on to check whether the candidate nearest node was a valid neighbor of the node from the previous frame (or if the candidate node was the same as the previous node). If so, then we set the current frame to be associated with the candidate node. If not, then the current and following frames were not labeled until a frame with a valid next node was found. Once a valid next node was found, the annotation tagging would resume from that frame and continue on until the end of the trajectory sequence.

After the node tagging process, the frames were annotated with edge labels. Suppose the frames at times $t_i, \ldots, t_{i+n}$ were labeled to be node $A$ and the frame at $t_{i+n+1}$ as node $B$. In this case, the frames at times $t_i, \ldots, t_{i+n}$ were labeled with the edge connecting node $A$ to node $B$.



\section{Navigation Plan Statistics}

\begin{figure*}
    \begin{subfigure}{0.19\linewidth}
    \includegraphics[width=\linewidth]{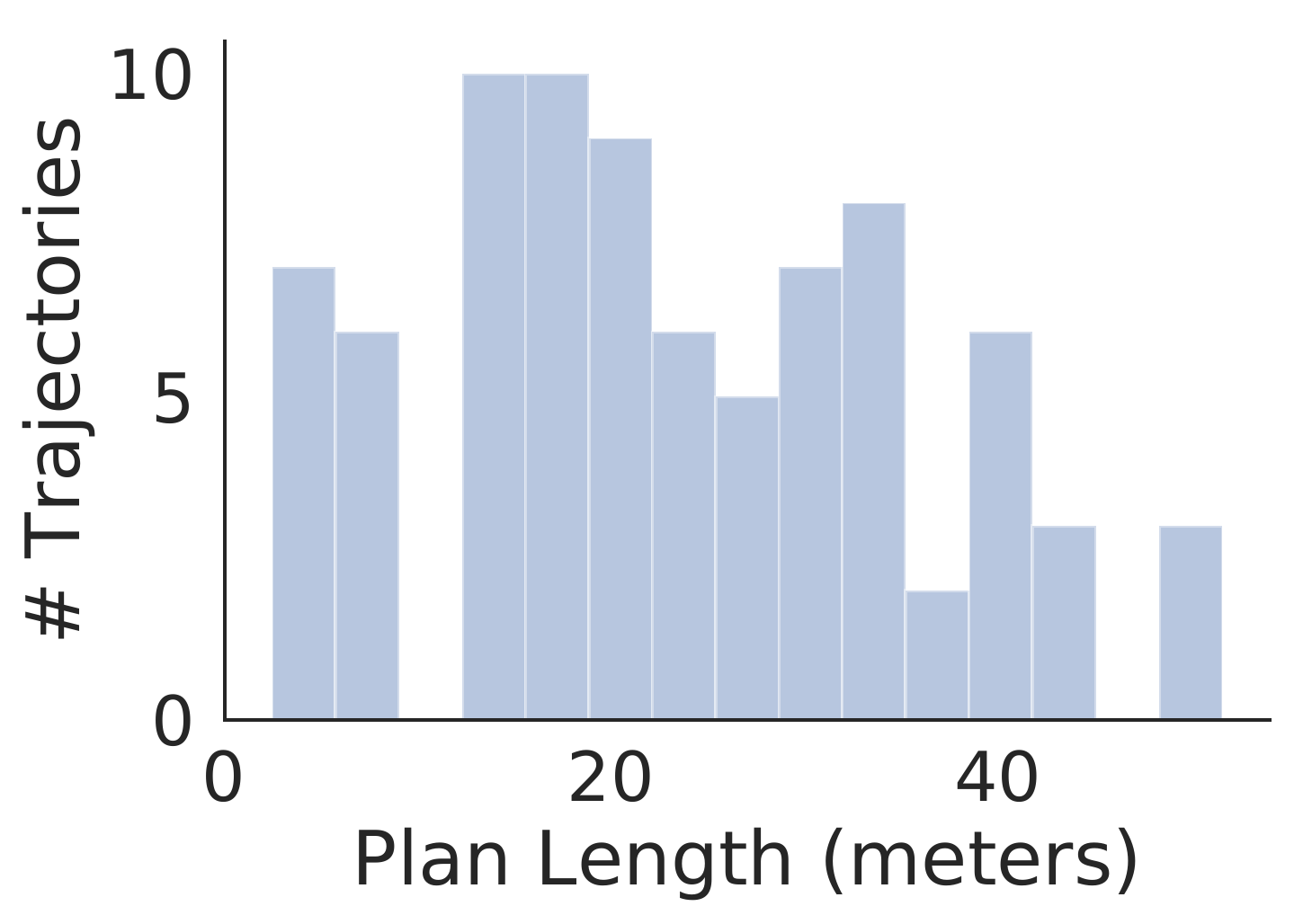}
    \caption{Area 1}
    \end{subfigure}
    \begin{subfigure}{0.19\linewidth}
    \includegraphics[width=\linewidth]{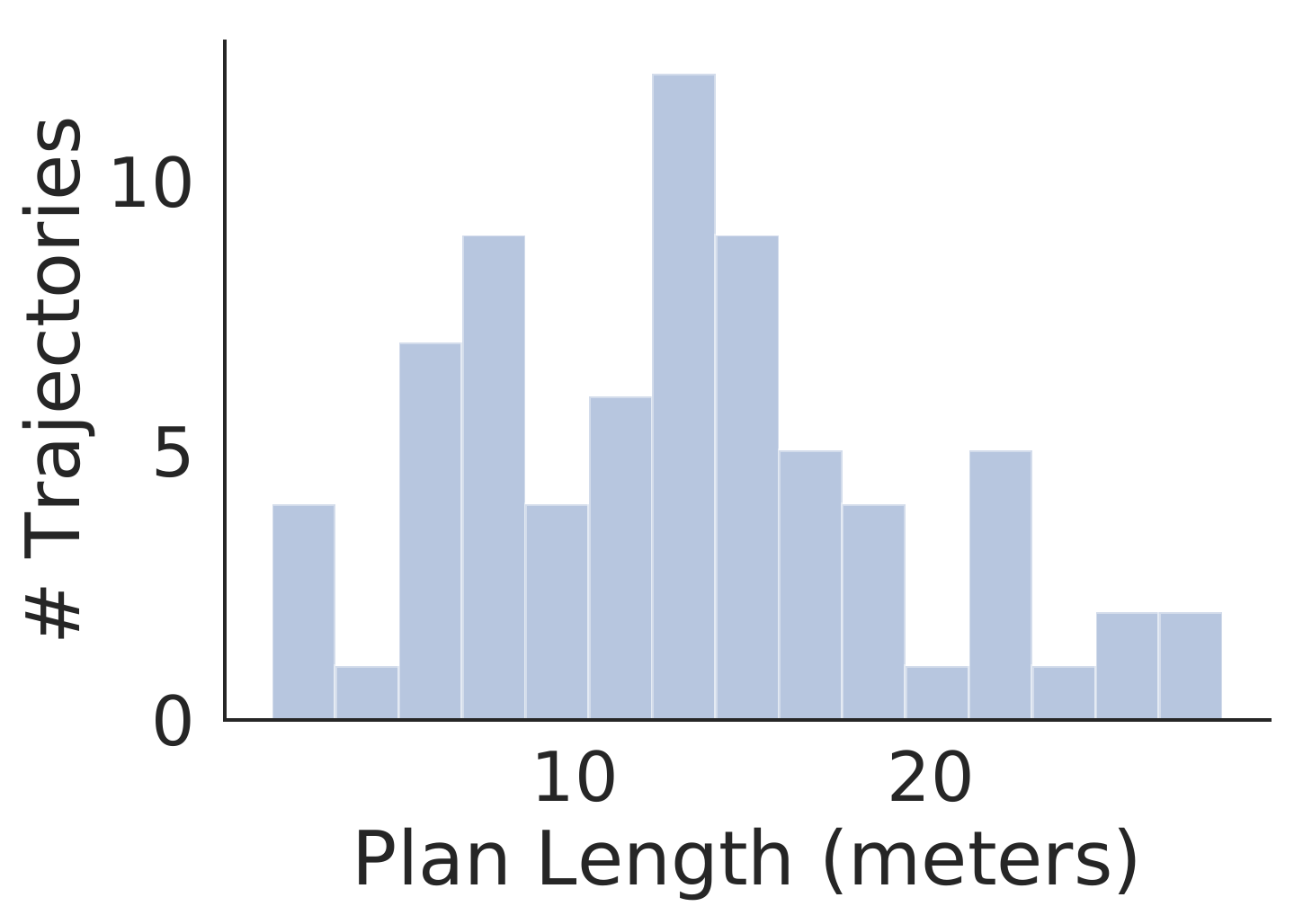}
    \caption{Area 3}
    \end{subfigure}
    \begin{subfigure}{0.19\linewidth}
    \includegraphics[width=\linewidth]{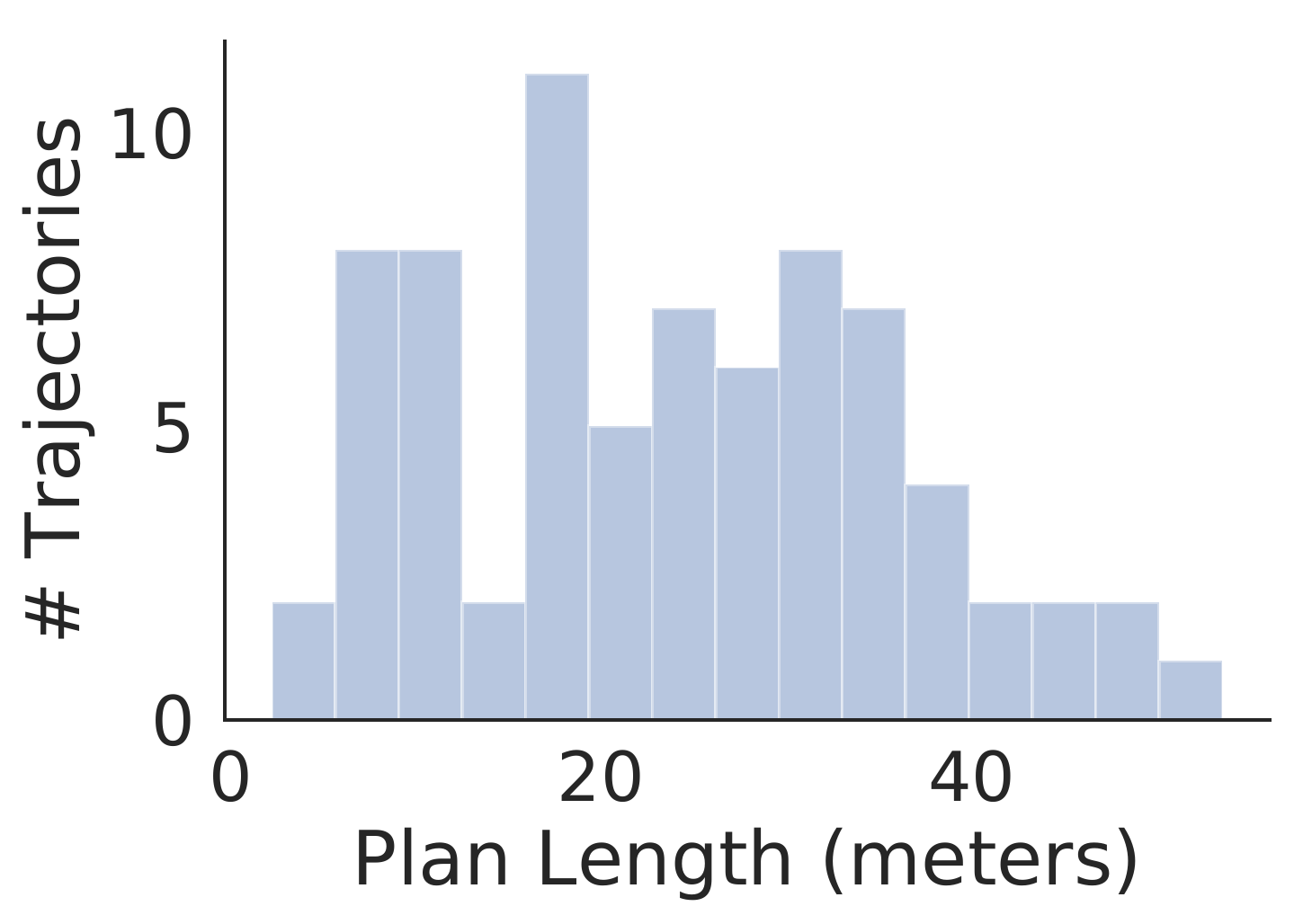}
    \caption{Area 4}
    \end{subfigure}
    \begin{subfigure}{0.19\linewidth}
    \includegraphics[width=\linewidth]{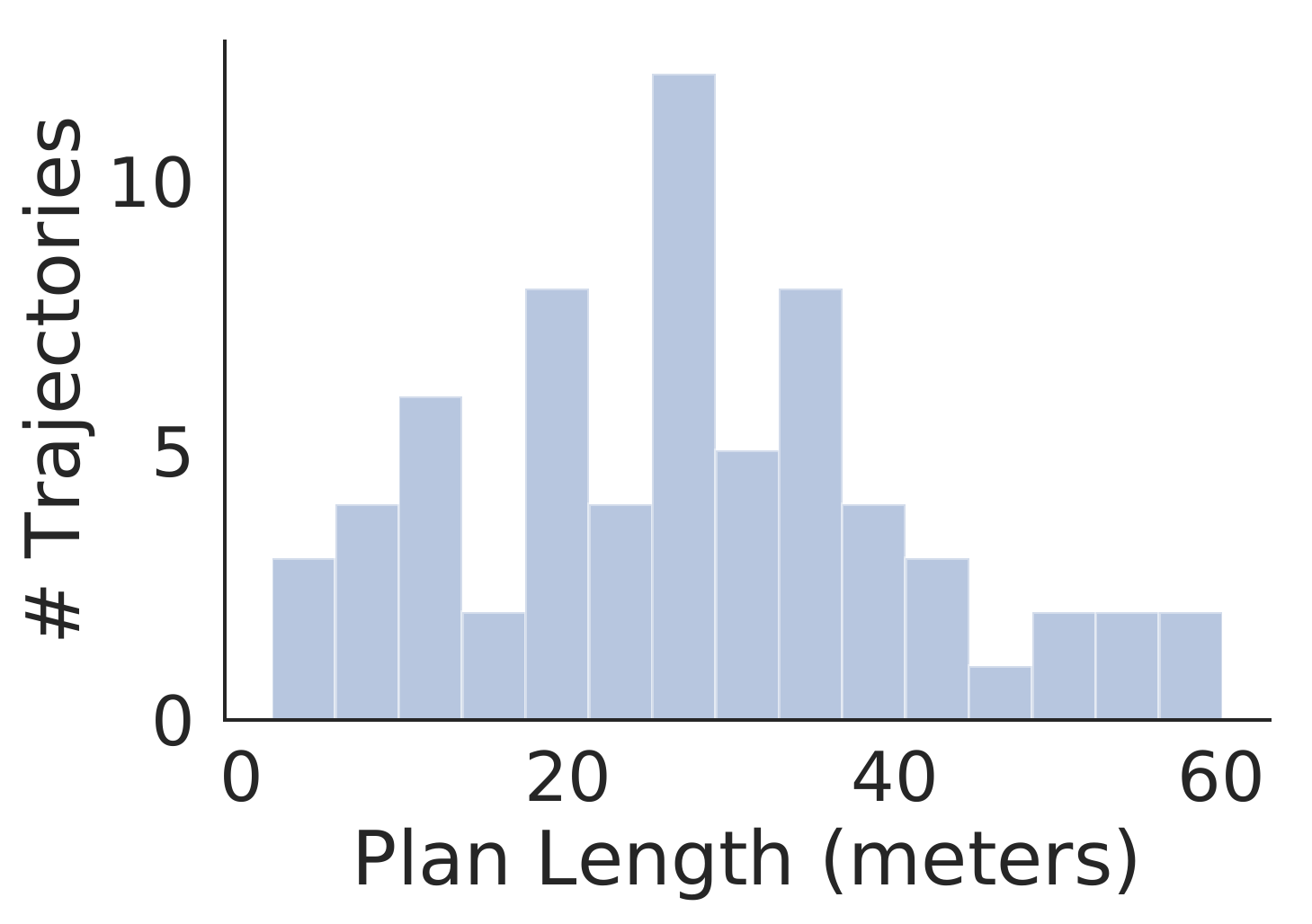}
    \caption{Area 5}
    \end{subfigure}
    \begin{subfigure}{0.19\linewidth}
    \includegraphics[width=\linewidth]{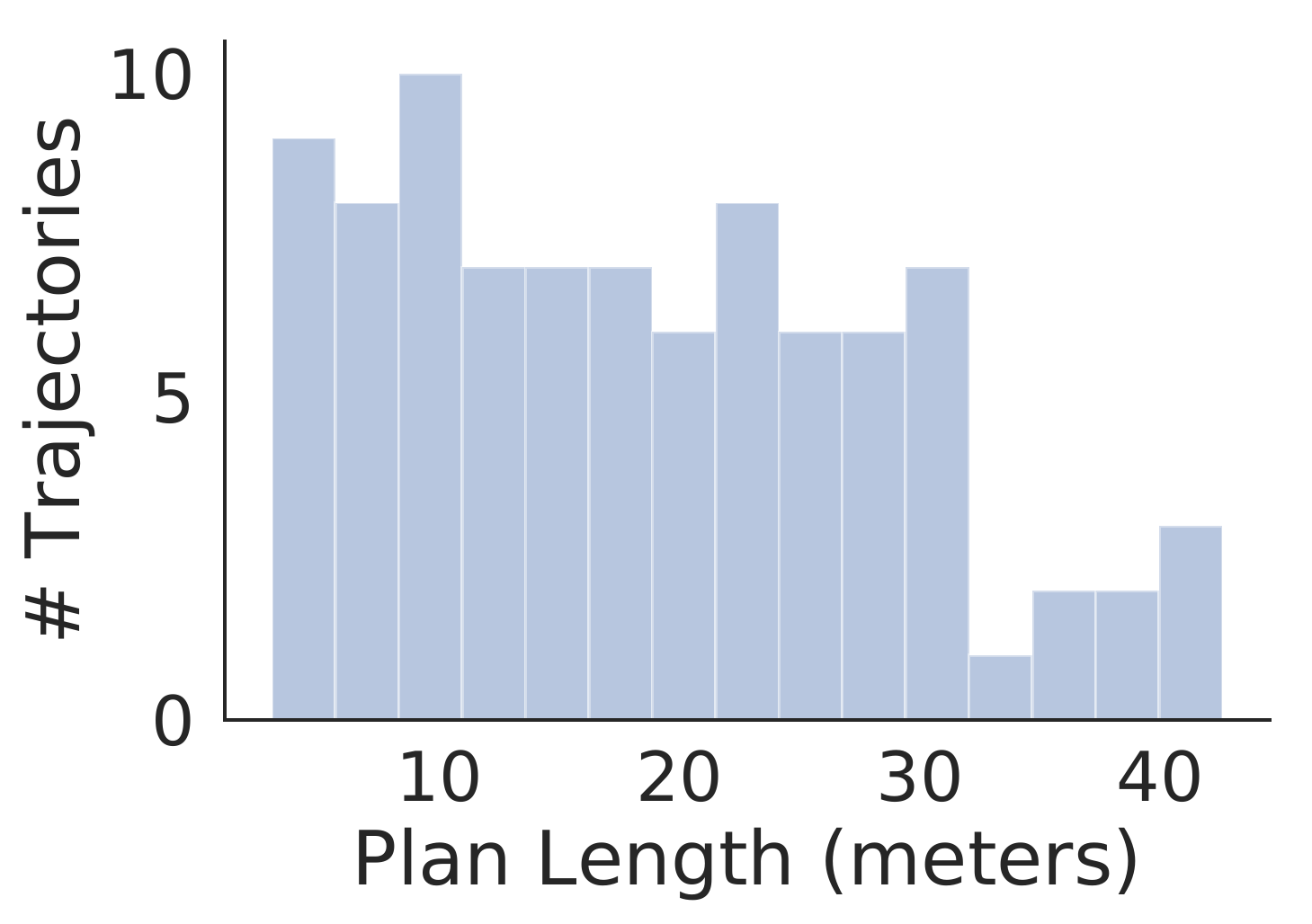}
    \caption{Area 6}
    \end{subfigure}
    \caption{Histogram of navigation plan lengths (meters) used for evaluation in each area of the Stanford 2D-3D-S dataset \cite{armeni_cvpr16,armeni_arxiv_2d3ds}.}
    \label{fig:path-length-statistics-meter}
\end{figure*}

\begin{figure*}
    \begin{subfigure}{0.19\linewidth}
    \includegraphics[width=\linewidth]{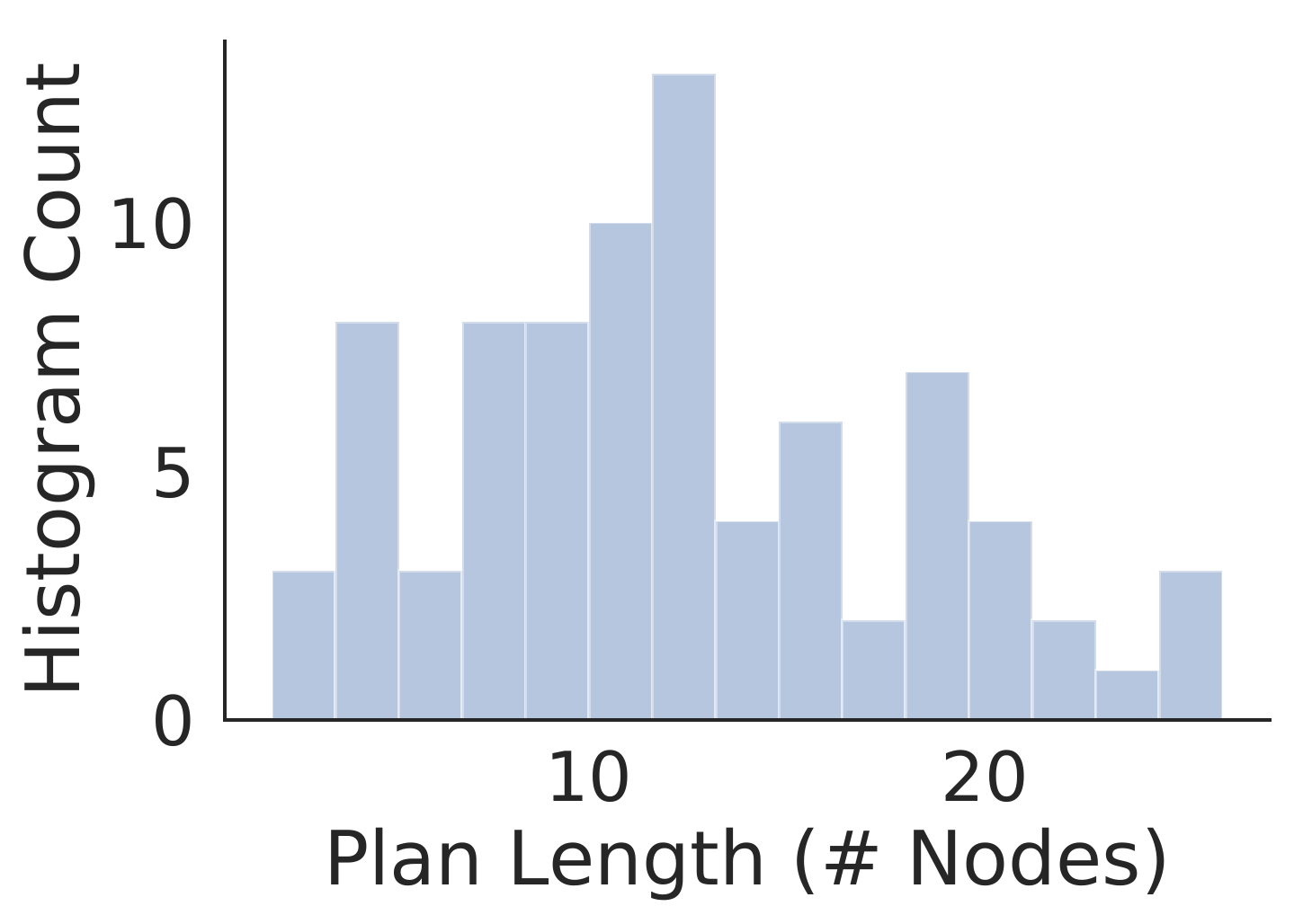}
    \caption{Area 1}
    \end{subfigure}
    \begin{subfigure}{0.19\linewidth}
    \includegraphics[width=\linewidth]{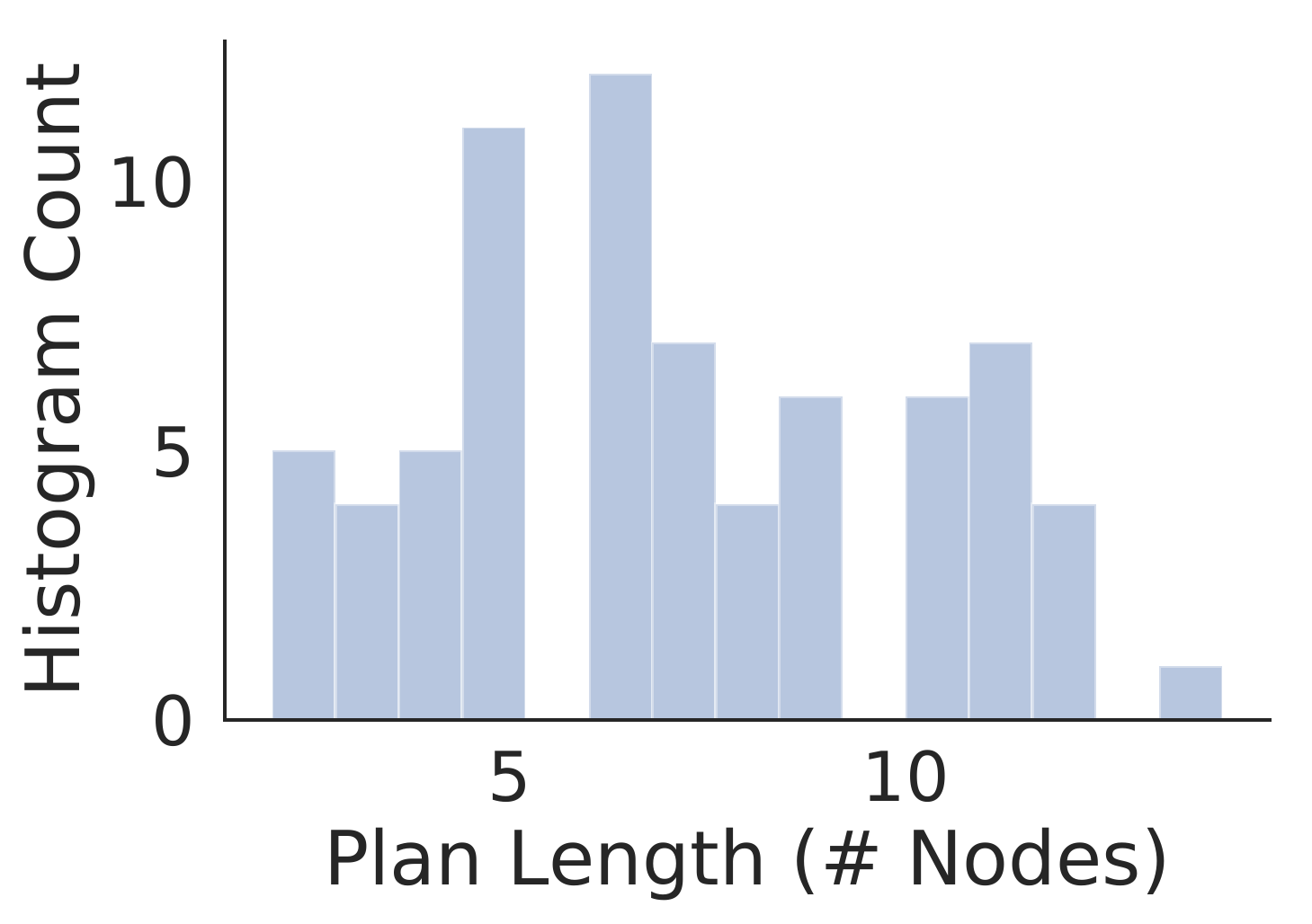}
    \caption{Area 3}
    \end{subfigure}
    \begin{subfigure}{0.19\linewidth}
    \includegraphics[width=\linewidth]{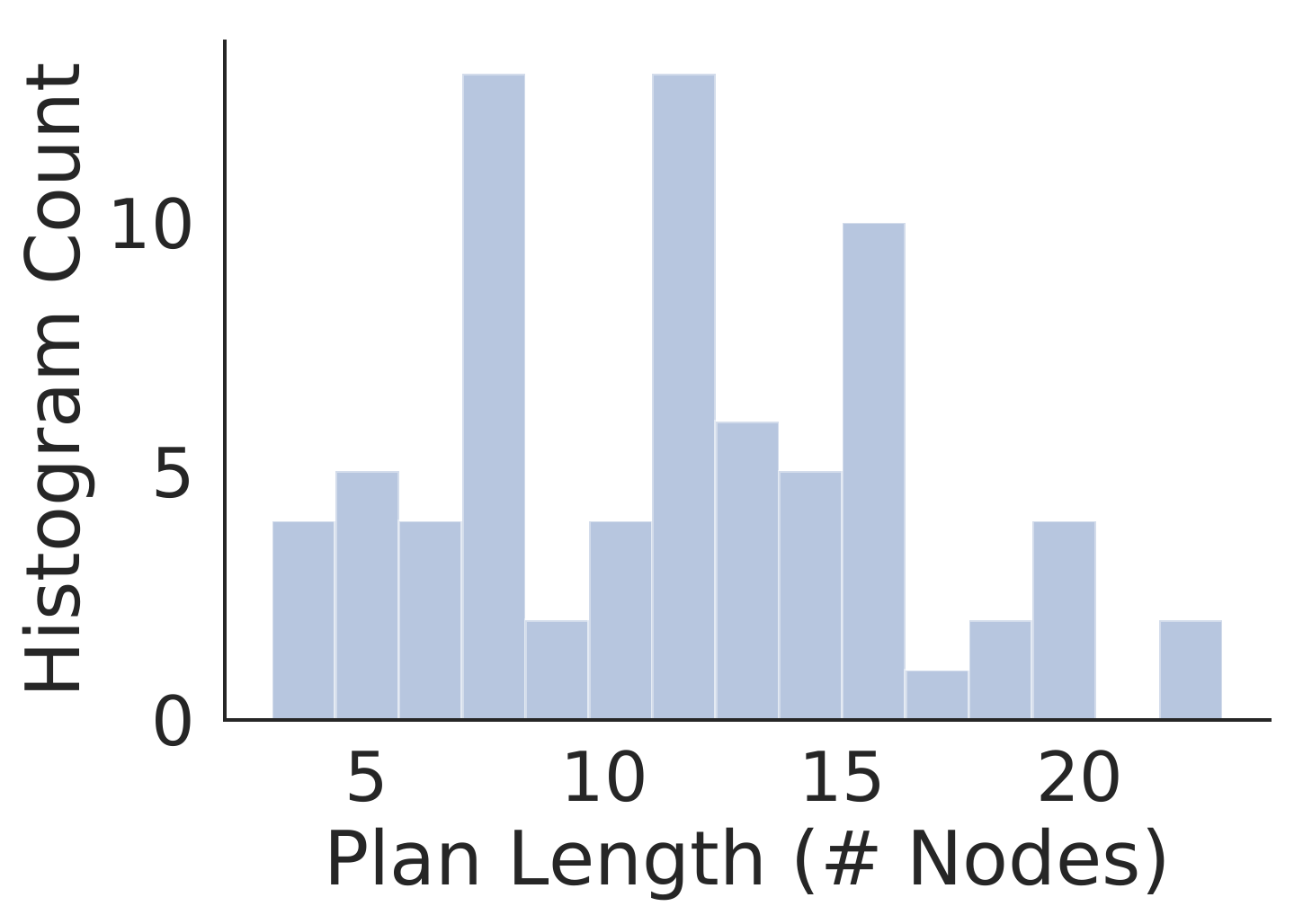}
    \caption{Area 4}
    \end{subfigure}
    \begin{subfigure}{0.19\linewidth}
    \includegraphics[width=\linewidth]{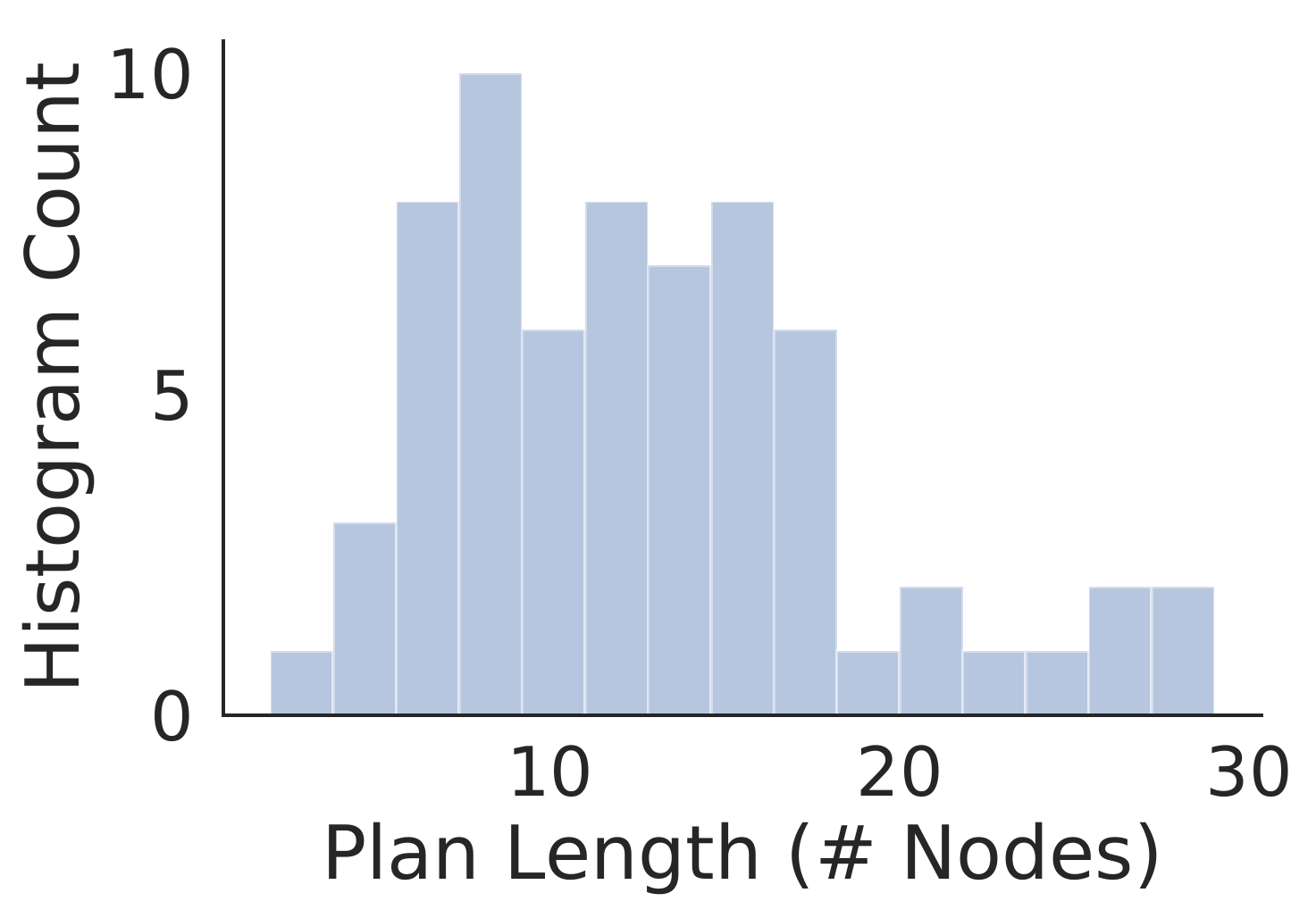}
    \caption{Area 5}
    \end{subfigure}
    \begin{subfigure}{0.19\linewidth}
    \includegraphics[width=\linewidth]{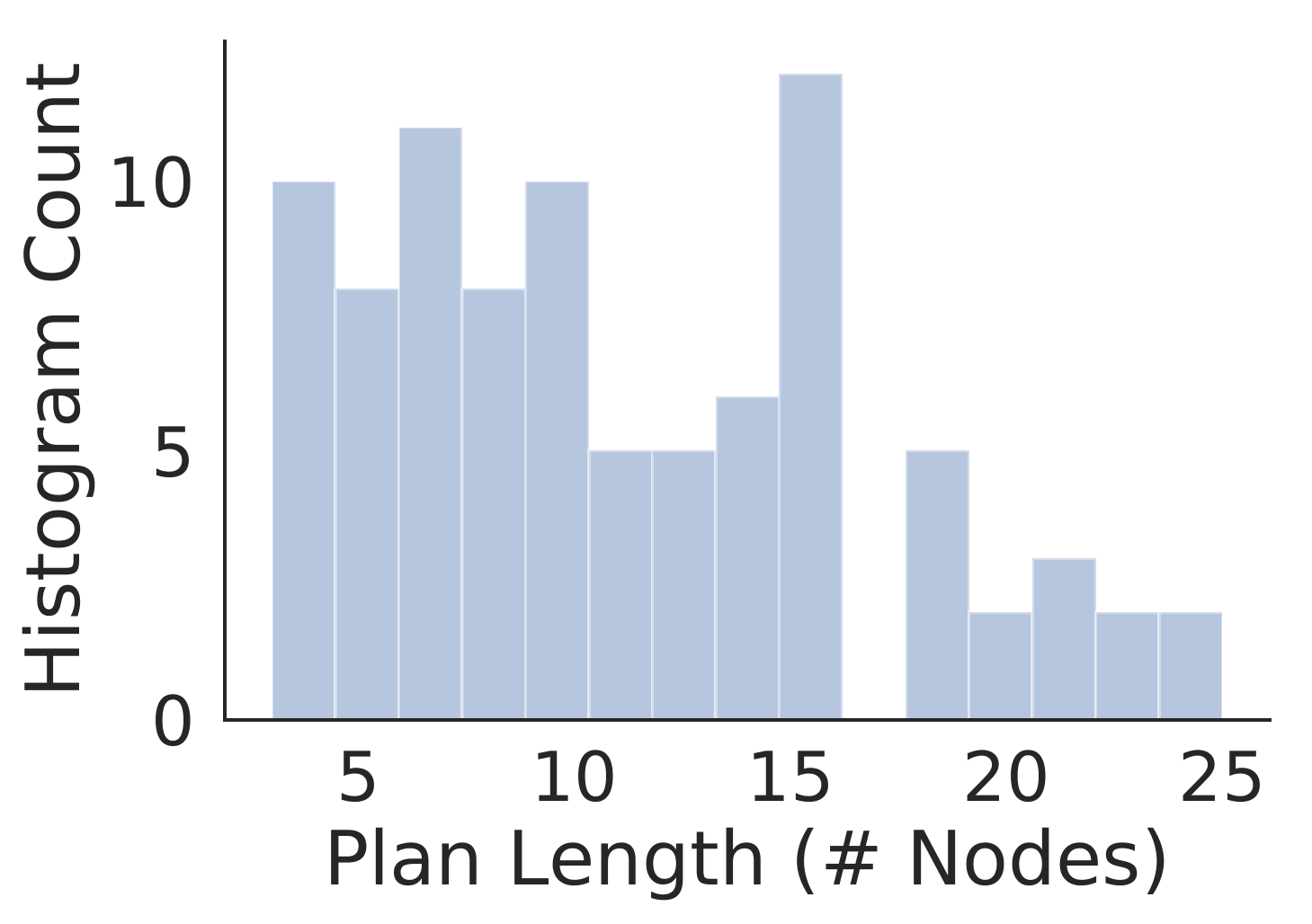}
    \caption{Area 6}
    \end{subfigure}
    \caption{Histogram of navigation plan lengths (\# nodes) used for evaluation in each area of the Stanford 2D-3D-S dataset \cite{armeni_cvpr16,armeni_arxiv_2d3ds}.}
    \label{fig:path-length-statistics-node}
\end{figure*}

To compare our approach against other baselines, we sampled navigation plans for each Stanford 2D-3D-S environment and tested different models with them. Each navigation plan was categorized into one of three different difficulties depending on the number of nodes in the navigation plan. A plan was in difficulty I if it contained 10 nodes or fewer, difficulty II if it contained 11 to 20 nodes, and difficulty III if it had more than 20 nodes. We provide an illustration of the distribution of plan lengths (\# nodes) for each area in Fig.~\ref{fig:path-length-statistics-node}. In particular, we can compare this with the histogram of path lengths measured in meters (Fig.~\ref{fig:path-length-statistics-meter}) to see how the number of nodes correlates with metric distance. For example, Area 5 has the longest plan lengths measured both in meters (60 m) and number of nodes (almost 30). Thus, it can be seen that the higher difficulty trajectories also tend to be over longer metric distances.

Interestingly, the results in the main paper show that our model performed the best on Area 5, which is the largest area, with path lengths going up to 60 m. The floorplan layout of area 5 is well structured, with spread out doorways and few open spaces, making it easier to navigate than the other areas.

\section{Additional Experimental Results}

\begin{figure*}
    \centering
    \includegraphics[width=\linewidth]{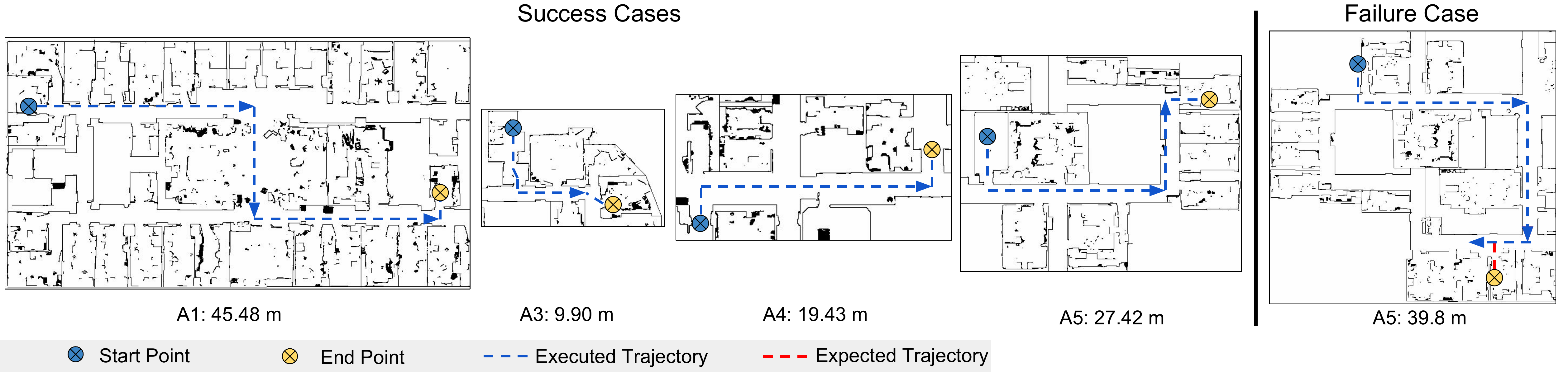}
    \caption{Examples of executed trajectories with the approximate navigation plan lengths. Seen environments: Areas 1, 5, 6 (A1, A5, A6). Unseen environments: Areas 3, 4 (A3, A4). Best viewed in color.}
    \label{fig:qualitative-trajectory-examples}
\end{figure*}

We show additional qualitative results in Fig.~\ref{fig:qualitative-trajectory-examples} and quantitative results in Table~\ref{tab:behavior-pred-results}. Table~\ref{tab:behavior-pred-results} shows the behavior prediction performance for each model by comparing the behavior prediction accuracy with the GTL model. More specifically, during the execution of each navigation model, we ran the node localizer in the background using the ground truth odometry of the robot. This allowed us to compare the agent's current node with nodes in the planned trajectory path to find the current ground truth behavior to execute. The reported behavior prediction accuracy is the accuracy of the model's selected behavior compared with the ground truth behavior coming from this lookup operation. This allows us to compare our GraphNav model against the baselines with less influence from the behavior network performance.

From Table~\ref{tab:behavior-pred-results}, we see that the PhaseNet and BehavRNN baseline approaches performed fairly poorly even on the areas that were seen during training. For example, the PhaseNet accuracies for the turn behavior were below 20\% on the train areas, whereas the GraphNav model achieved 40\% to 70\% accuracy. Surprisingly, the BehavRNN had accuracies of less than 20\% for the \textit{corridor follow} behavior, which is the most commonly used behavior.

In the unseen environments, the performance of the PhaseNet and BehavRNN baselines remained fairly similar compared to the performance in seen environments. For the GraphNav model, the turn behavior prediction accuracies were lower than in the seen areas, which is unsurprising since the agent had not seen the area before. Thus, the localization network struggled more with the unseen spaces compared to the seen spaces. However, the \textit{find door} performance still remained very high. This indicates that the \textit{find door} behavior is robust even in previously unseen environments, and that the localization network is very accurate at detecting when the agent is located within the room.

Lastly, we report results for the graph localization network prediction accuracy in Table~\ref{tab:localization-accuracy-results}. The accuracy in the train areas was very high at 89.8\%, whereas the validation and test area accuracies were substantially lower. One way to reduce this gap between train and test performance, and to prevent overfitting, is by diversifying the training dataset with more environments. Currently, there are only three environments in the training set, two of which are very similar in structure since they come from two floors of the same building. Training with more data from different spaces would ideally let the model generalize better to unseen environments.

Nevertheless, our navigation system still works well and is able to navigate in both seen and unseen environments, outperforming the relevant baseline approaches. As shown in Fig.~\ref{fig:qualitative-trajectory-examples} as well as in the supplementary video, our navigator can traverse through complex, cluttered environments and reach the destination without colliding. Given these results for visual navigation using topological maps, we believe that this is a promising direction of research and hope that our work inspires further research in autonomous visual navigation.
\begin{table}
    \centering
    \begin{tabular}{lr}
        \toprule
        Split & Accuracy \\
        \midrule
        Train (Areas 1, 5, 6) & 89.8 \\
        Validation (Area 3) & 52.3 \\
        Test (Area 4) & 31.1 \\
        \bottomrule
    \end{tabular}
    \caption{The localization prediction accuracies from the graph localization network on trajectories from our dataset.}
    \label{tab:localization-accuracy-results}
\end{table}

\begin{table*}
    \centering
    \begin{tabular}{ccccccc}
        \toprule
         & & \multicolumn{5}{c}{Per-Behavior Prediction Accuracies} \\
        \cmidrule(l){3-7}
        Area ID & Model & cf & fd & tl & tr & s \\
        \midrule
        1 (Seen) & PhaseNet & 57.6 (8596) & 61.3 (2044) & 14.6 (1536) & 14.0 (1467) & \textbf{100} (56) \\
        5 (Seen) & PhaseNet & 39.1 (8217) & 21.4 (7704) & 17.5 (2407) & 10.8 (1295) & - (0) \\
        6 (Seen) & PhaseNet & 70.3 (5494) & 88.3 (2030) & 16.1 (1796) & 16.5 (1467) & \textbf{100} (56) \\
        1 (Seen) & BehavRNN & 16.3 (1716) & 77.1 (952) & 21.6 (361) & 21.0 (391) & 0 (22) \\
        5 (Seen) & BehavRNN & 13.2 (1740) & 64.1 (1261) & 33.4 (479) & 32.8 (408) & - (0) \\
        6 (Seen) & BehavRNN & 6.9 (1572) & 75.1 (995) & 28.4 (500) & 42.2 (412) & 0 (7) \\
        1 (Seen) & GraphNav (ours) & 89.5 (15164) & 91.9 (2026) & 69.4 (2537) & \textbf{76.0} (2066) & 52.9 (221) \\
        5 (Seen) & GraphNav (ours) & 89.6 (16499) & 88.9 (2155) & 46.6 (2732) & 51.2 (2558) & 50.5 (103) \\
        6 (Seen) & GraphNav (ours) & 89.6 (13094) & 96.0 (2063) & 68.7 (2710) & 58.3 (3018) & 51.5 (101) \\
        1 (Seen) & GraphNavPF (ours) & \textbf{95.9} (15607) & \textbf{92.4} (2011) & \textbf{72.6} (2434) & 75.6 (2027) & 54.3 (208) \\
        5 (Seen) & GraphNavPF (ours) & \textbf{96.7} (19981) & \textbf{95.0} (2146) & \textbf{56.5} (3375) & \textbf{66.4} (2870) & \textbf{60.3} (252) \\
        6 (Seen) & GraphNavPF (ours) & \textbf{96.1} (14508) & \textbf{96.8} (2001) & \textbf{75.0} (3105) & \textbf{73.0} (3241) & 67.9 (112) \\
        1 (Seen) & GTL\textsuperscript{\textdagger} & 100 (16898) & 100 (1965)  & 100 (2731) & 100 (2175) & 100 (137) \\
        5 (Seen) & GTL\textsuperscript{\textdagger} & 100 (24194) & 100 (2210) & 100 (3585) & 100 (3241) & 100 (303) \\
        6 (Seen) & GTL\textsuperscript{\textdagger} & 100 (15005) & 100 (2061) & 100 (3006) & 100 (3641) & 100 (126) \\
        \midrule
        3 (Unseen) & PhaseNet & 57.0 (4614) & 92.6 (1305) & 15.0 (1425) & 19.9 (2051) & \textbf{51.6} (64) \\
        3 (Unseen) & BehavRNN & 14.8 (2118) & 62.4 (681) & 55.2 (630) & 39.3 (732) & 0 (13) \\
        3 (Unseen) & GraphNav (ours) & 78.8 (7208) & 88.9 (1374) & 36.3 (1811) &  25.0 (2576) & 28.3 (106) \\
        3 (Unseen) & GraphNavPF (ours) & \textbf{85.7} (8552) & \textbf{99.2} (1314) & \textbf{44.0} (2231) & \textbf{40.9} (2534) & 50.0 (122) \\
        3 (Unseen) & GTL\textsuperscript{\textdagger} & 100 (9214) & 100 (1468) & 100 (2100) & 100 (2462) & 100 (130) \\
        \midrule
        4 (Unseen) & PhaseNet & 62.6 (8200) & 50.9 (1395) & 15.0 (1545) & 23.2 (2083) & - (0) \\
        4 (Unseen) & BehavRNN & 16.9 (1792) & 66.4 (657) & \textbf{33.5} (382) & 21.0 (509) & - (0) \\
        4 (Unseen) & GraphNav (ours) & \textbf{76.2} (10793) & \textbf{91.1} (1486) & 31.0 (2169) & 21.6 (2245) & - (0) \\
        4 (Unseen) & GraphNavPF (ours) & 73.6 (9781) & 89.6 (1417) & 32.9 (2169) & \textbf{37.3} (2092) & - (0) \\
        4 (Unseen) & GTL\textsuperscript{\textdagger} & 100 (16554) & 100 (1347) & 100 (2888) & 100 (2819) & - (0) \\
        \bottomrule
    \end{tabular}
    \caption{Performance comparison using success rate (SR) and average plan completion (PC). Numbers in parentheses represent total number of attempts for that entry. The \textsuperscript{\textdagger} indicates that GTL utilizes additional ground truth information.}
    \label{tab:behavior-pred-results}
\end{table*}
\end{appendices}

\end{document}